\documentclass[journal]{IEEEtran}

\usepackage[utf8]{inputenc}
\usepackage{mathrsfs}

\usepackage[dvipsnames,table,xcdraw]{xcolor}

\usepackage{amsmath,amsfonts,amssymb,mathtools,bbm,dsfont,mathdots}
\usepackage{amsthm}

\usepackage{tikz}
\usepackage{tikz-cd}
\usetikzlibrary{shapes.geometric, arrows.meta, arrows}

\usepackage{algorithm2e}
\SetKwComment{Comment}{/* }{ */}
\SetKwProg{Init}{Initialize}{}{}
\RestyleAlgo{ruled}

\usepackage[caption=false,font=footnotesize]{subfig}
 
\usepackage{graphicx,float}
\usepackage{booktabs,multirow} % Removed duplicate booktabs

\usepackage[colorlinks=true, pdfborder={0 0 0}, linkcolor=red]{hyperref}
\allowdisplaybreaks
\def\E{\mathbb{E}}
\def\N{\mathbb{N}} 
\renewcommand{\P}{\mathbb{P}} % Changed to renewcommand

\def\R{\mathbb{R}}
  
\renewcommand{\L}{\mathcal{L}} % Changed to renewcommand 
 
\def\1{\mathbbm{1}}

\def\II{\text{II}}

\newcommand{\ind}{\mathds{1}}
\newcommand{\Pbb}{\mathbb{P}}
\newcommand{\db}[1]{\boldsymbol{#1}}

\newtheorem{assumption}{Assumption}
\newtheorem{theorem}{Theorem}
\newtheorem{proposition}{Proposition}
\newtheorem{corollary}{Corollary}
\newtheorem{lemma}{Lemma}
\newtheorem{remark}{Remark}

\newtheorem{conjecture}{Conjecture}

\begin{document}

\title{Central Limit Theorems for Stochastic Gradient Descent Quantile Estimators}

\author{Ziyang~Wei,
        Jiaqi~Li, 
        Likai~Chen,
        and~Wei~Biao~Wu% 
\thanks{Z. Wei, J. Li, and W. B. Wu are with the Department of Statistics, University of Chicago (e-mail: jqli@uchicago.edu). (Corresponding author: Jiaqi Li.)}% 
\thanks{L. Chen is with the Department of Statistics and Data Science, Washington University in St. Louis.}%
\thanks{Manuscript received May 29, 2025; revised January 8, 2026.}}

% The paper headers
\markboth{IEEE Transactions on Information Theory,~Vol.~X, No.~X, June~2026}%
{Wei \MakeLowercase{\textit{et al.}}: Central Limit Theorems for Stochastic Gradient Descent Quantile Estimators}

\maketitle

\begin{abstract}
This paper develops asymptotic theory for quantile estimation via stochastic gradient descent (SGD) with a constant learning rate. The quantile loss function is neither smooth nor strongly convex. Beyond conventional perspectives and techniques, we view quantile SGD iteration as an irreducible, periodic, and positive recurrent Markov chain, which cyclically converges to its unique stationary distribution regardless of the arbitrarily fixed initialization. To derive the exact form of the stationary distribution, we analyze the structure of its characteristic function by exploiting the stationary equation. We also derive tight bounds for its moment generating function (MGF) and tail probabilities. Synthesizing the aforementioned approaches, we prove that the centered and standardized stationary distribution converges to a Gaussian distribution as the learning rate $\eta\rightarrow0$. This finding provides the first central limit theorem (CLT)-type theoretical guarantees for the quantile SGD estimator with constant learning rates. We further propose a recursive algorithm to construct confidence intervals of the estimators with statistical guarantees. Numerical studies demonstrate the effective finite-sample performance of the online estimator and inference procedure. The theoretical tools developed in this study are of independent interest for investigating general SGD algorithms formulated as Markov chains, particularly in non-strongly convex and non-smooth settings.
\end{abstract}

\begin{IEEEkeywords}
stochastic gradient descent, statistical inference, quantile estimation, asymptotic normality
\end{IEEEkeywords}

\IEEEpeerreviewmaketitle

\section{Introduction}
\IEEEPARstart{A} fundamental task in statistical learning is parameter estimation via the minimization of an objective function. The rapid collection of increasingly massive datasets has exposed the limitations of classical full-batch optimization methods. Stochastic Gradient Descent (SGD), also known as the Robbins-Monro algorithm (\cite{robbins1951stochastic}), has emerged as a leading approach to address this computational issue. With a diverse range of variations and modifications (\cite{polyak_acceleration_1992,shamir2013stochastic,kingma_adam_2014,li_asymptotics_2024,zhong_probabilistic_2024}), it has become a standard tool in machine learning and artificial intelligence. The computation and storage efficiency due to the recursive nature of SGD make it well-suited for streaming data and sequential learning tasks at scale. The statistical inference for stochastic approximation methods under smooth and strongly convex conditions has been systematically investigated (\cite{Lij_2024}). In their seminal works, \cite{polyak_acceleration_1992} and \cite{pflug_stochastic_1986} established the asymptotic normality of averaged SGD with the decaying learning rate (step size) and the last iterate of SGD with the constant learning rate. 

This paper focuses on the online quantile estimation and inference with SGD, a single-pass algorithm. Quantile estimation and regression have significant and broad applications across various fields. Quantiles serve as more robust location parameters than the expectation since they are less susceptible to heavy-tailed distributions and outliers. Moreover, they offer a holistic and detailed perspective of the target distribution, allowing practitioners to tailor the model to their risk preferences and specific goals.  

Traditional quantile estimators based on order statistics have well-established large-sample properties, as studied by \cite{bahadur1966note} and \cite{kiefer1967bahadur}. However, these methods are computationally inefficient for handling large-scale, sequentially arriving data due to their high memory demands. Online quantile estimation and inference have gained growing interest in recent years (\cite{luo_quantiles_2016, dzhamtyrova2020competitive, ichinose2023online,chen2024renewable}). Recent works, such as \cite{volgushev_distributed_2019} and \cite{Chen18lwd}, introduced novel algorithms for conditional quantile estimation that address computational and memory challenges. The obstacles to the theoretical study of the quantile estimation and regression problem come from its non-smoothness and lack of strong convexity. Consequently, a majority of existing approaches and results for stochastic approximation theory become inapplicable.

The asymptotic normality of the averaged stochastic gradient descent (ASGD) solution to quantile estimation with decaying learning rate was shown by \cite{Bardou2009}. In \cite{cardot2013efficient, cardot2017online}, the authors studied the non-asymptotic behavior and uncertainty quantification of ASGD in the context of geometric median estimation for multivariate distributions. \cite{costa2021non,chen2023recursive} further analyzed the finite sample performance of online quantile estimation by establishing upper bounds for the $L^q$ moments, the moment generating function, and the tail probability of SGD and ASGD. In terms of recursive quantile regression, \cite{shen2025online} designed a mixed step size schedule SGD with high-probability error bounds, and \cite{lee_fast_2025} proposed a random scaling procedure that enables fast inference for ASGD. Despite these advances, existing literatures have primarily focused on stochastic approximation methods for online quantile estimators with decaying learning rates, which introduces additional tuning parameters and complicates practical implementation. In contrast, constant learning-rate schemes have recently gained popularity due to easy parameter tuning and robust empirical performance. Investigating constant learning-rate SGD for quantile estimation is particularly important, as practical applications often rely on parallelizing multiple SGD sequences for faster convergence and use the extrapolation techniques to de-bias the SGD estimator. Moreover, deriving theoretical results under constant learning rates is, surprisingly, mathematically more challenging due to non-diminishing step-sizes, requiring more sophisticated analysis (\cite{cardot2013efficient, cardot2017online}). Recently, \cite{zhang_piecewise_2025} applied a clever piecewise Lyapunov function approach and obtained moment bounds for SGD iterates with sub-quadratic loss functions. However, The existence of a stationary distribution for quantile SGD iterates under a constant learning rate remains unexplored, as does the formal derivation of a corresponding weak convergence theory. In this paper, we provide a partial solution to this open problem by investigating SGD with quantiles being rational numbers. We also propose a conjecture for the more challenging irrational cases (cf. Conjecture~\ref{conjecture_irrational}) and leave it for future study.

\subsection{Our Contributions}
Suppose that we have sequentially arriving i.i.d samples $X_1,X_2,\ldots$ with cumulative distribution $F(x)=\P(X \leq x)$. Given a quantile level $\tau \in (0,1)$, we aim to estimate the $\tau$-th quantile of the distribution, defined as the optimizer of the quantile loss function:
\begin{align}
\label{eq_minimizer}
\theta(\tau)=\mathop{\mathrm{arg\,min}}_{\theta \in \mathbb{R}} \E\{(X-\theta)(\tau - \mathbbm{1}_{\theta\ge X})\}.
\end{align}
Consider the constant learning-rate SGD algorithm that iteratively updates the values of the estimator
\begin{align}\label{sgd}
    \theta_{n+1}(\tau)=\theta_{n}(\tau)+\eta [ \tau \mathds{1}_{X_{n+1}>\theta_{n}(\tau)} -(1-\tau) \mathds{1}_{X_{n+1}\leq \theta_{n}(\tau)} ],
\end{align}
where $\eta >0$ is the fixed learning rate. Let $\theta_{\infty}(\tau)$ denote the random variable following the stationary distribution of the Markov chain induced by \eqref{sgd}. The diagram below illustrates the key ingredients of our analysis:
$$
\dfrac{\theta_n(\tau) - \theta(\tau)}{\sqrt{\eta}} 
\xrightarrow[n \rightarrow \infty]{\mathcal{D}} 
\dfrac{\theta_{\infty}(\tau) - \theta(\tau)}{\sqrt{\eta}} 
\xrightarrow[\eta \rightarrow 0]{\mathcal{D}} 
\mathcal{N}(0, V),
$$
where $V$ is the limiting variance of $(\theta_{\infty}(\tau) - \theta(\tau))/\sqrt{\eta}$ which will be specified later. In the literature, for SGD quantile estimation with fixed learning rates, both the convergence to the stationary measure and the convergence to the normality have not been discussed.

Our contribution in this paper is three-fold. \textbf{(a)} We first leverage Foster's lemma (see, e.g., \cite{Markov,Bremaud1999}) to demonstrate that the constant learning-rate SGD of quantile loss forms a positive recurrent Markov chain. Hence, it has a unique stationary distribution (Section \ref{setup}). \textbf{(b)} To further investigate its asymptotic properties, we invoke the technique developed by \cite{tweedie1983existence} to bound the moment generating function of the stationary distribution, as well as its first and second derivatives. It enables us to control the tail behavior of the stationary probability and its first and second moments (Section \ref{Theory}). \textbf{(c)} Combining these prerequisites, we achieve the important conclusion on the asymptotic normality of SGD iterates for quantile loss functions, which facilitates an online inference method for the SGD estimator. In Section \ref{simulation}, we conduct numerical studies that demonstrate our theoretical results, including estimation and inference of quantiles with satisfactory finite sample performance. Detailed proofs and some extensions are discussed at the Appendix of the paper.

\subsection{Related Works}
\textbf{Asymptotics of SGD.} The asymptotic behavior of SGD has been extensively studied. Early foundational work by \cite{blum_approximation_1954, dvoretzky_stochastic_1956, sacks_asymptotic_1958} established conditions for convergence of SGD iterates to a minimizer of the objective function. Subsequent research refined these results by providing stronger theoretical guarantees, such as almost sure convergence (\cite{fabian_asymptotic_1968, robbins_convergence_1971, ljung_analysis_1977, lai_stochastic_2003}). A key perspective in the analysis of constant learning-rate SGD is viewing it as a homogeneous Markov chain, enabling the study of its stationary distribution and long-run behavior. See for instance, \cite{pflug_stochastic_1986} studied the stationary solutions of constant learning-rate SGD, and \cite{dieuleveut_bridging_2020,merad2025convergence} demonstrated its convergence to a unique stationary distribution in the Wasserstein-2 distance. An alternative approach interprets SGD as an iterated random function, as explored in \cite{dubins_invariant_1966,barnsley_iterated_1985,diaconis_iterated_1999}, with applications in heavy-tailed stochastic optimization (\cite{mirek_heavy_2011,gupta_limit_2020,gupta_convergence_2021,gurbuzbalaban_heavy-tail_2021,hodgkinson_multiplicative_2021}). To investigate heavy-tailed noise settings (\cite{krasulina_stochastic_1969, buraczewski_asymptotics_2012,cuny_martingale_2014,wang_convergence_2021}), recent work by \cite{Lij_2024} has applied geometric moment contraction (GMC) techniques (\cite{wu_limit_2004}) to establish SGD convergence in the Euclidean norm, providing a more comprehensive asymptotic framework. However, most of the existing works on constant learning-rate stochastic approximation focused on i.i.d. noise sequences (\cite{chen2022stationary}) as well as strongly convex and smooth settings. For the works on general non-convex optimization, a dissipativity assumption is usually imposed (\cite{raginsky_non-convex_2017, erdogdu_global_2018, xu_global_2018, yu_analysis_2021}), which is not satisfied by the quantile loss function. 

\noindent \textbf{Quantile estimation.} Traditional quantile estimators based on order statistics have well-established large-sample properties (\cite{bahadur1966note,kiefer1967bahadur}), but they are inefficient for large-scale, sequential data due to high memory demands. Online quantile estimation (\cite{luo_quantiles_2016,dzhamtyrova2020competitive,ichinose2023online,chen2024renewable}) and inference (\cite{Chen18lwd,volgushev_distributed_2019,shen2025online}) have gained interest to address these issues, though most focus on asymptotic normality under decaying learning rates (\cite{cardot2013efficient,chen2023recursive}), which require additional tuning and complicate practical use (\cite{cardot2017online}). To bridge this gap, we propose to apply the constant learning-rate SGD algorithm to the quantile estimation and derive the stationary distribution of SGD estimators, enabling the study of stability in this challenging non-smooth and non-strongly-convex scenario.

\noindent\textbf{Learning rates.} Different learning rates have been adopted in the literature. See for example,  \cite{pflug_stochastic_1986,dieuleveut_bridging_2020,merad2025convergence,Huo_2026} researched on stationary solution under constant learning rate among researchers by interpreting the SGD process as a homogeneous Markov chain. For decreasing learning rates, \cite{rakhlin_making_2011} developed the optimal convergence rate with $\gamma_n=\gamma/n$; \cite{Ge_2019} showed the convergence property of polynomial decaying learning rate $\gamma_n=Cn^{-\beta}$ for some constant $C>0$ and $\beta\in(1/2,1)$ in both convex and non-convex cases. Moreover, \cite{gower_sgd_2019,nguyen_tight_2019} considered a burn-in regime with a constant learning rate $\gamma$ for the early stage and a decreasing learning rate $\gamma_n$ for a later stage. See also \cite{loizou_stochastic_2021,wang_convergence_2023,jiang_adaptive_2024} for a wide range of adaptive learning rates. We would like to emphasize that the theoretical properties of SGD with constant learning rates are more difficult to derive, as the existence and uniqueness of a stationary distribution are undiscussed in many settings. Even it exists, it is also nontrivial to characterize such distribution. We propose a novel method in this paper to address this gap.

\noindent\textbf{Online inference.} Beyond convergence analysis, online inference for SGD-type estimators is also critical, especially for uncertainty quantification. Traditional inference methods for M-estimators, such as bootstrap procedures \cite{fang_online_2018,fang_scalable_2019,zhong_online_2023}, are often impractical in online settings due to their high computational cost. An alternative approach involves leveraging the Polyak-Ruppert averaging technique (\cite{ruppert_efficient_1988,polyak_acceleration_1992}), which improves statistical efficiency and facilitates inference. The averaged SGD (ASGD) sequence (\cite{gyorfi_averaged_1996,defossez_averaged_2015}) has been shown to achieve asymptotic normality at an optimal convergence rate (\cite{moulines_non-asymptotic_2011, dieuleveut_nonparametric_2016, dieuleveut_harder_2017,jain_parallelizing_2018}). However, inference for the last iterate of constant learning-rate SGD is even more challenging and rarely discussed in the literature. We shall fill in this gap by providing the quenched CLT of the SGD quantile estimator as $\eta\rightarrow0$, regardless of the arbitrary initialization. Furthermore, online inference methods using blocking-based variance estimation (\cite{chen_statistical_2020,Zhu2023}) and recursive kernel estimation (\cite{huang_recursive_2014}) have been developed to achieve optimal mean squared error rates while accommodating dependence structures, enabling practical and theoretically sound online inference for SGD-based estimators.

\subsection{Notation}
We use $\mathbb{P}(A)$ to denote the probability of the event $A$. For a vector $v = (v_{1}, \ldots, v_{d})^{\top} \in\mathbb{R}^{d}$ and $q > 0$, we denote $|v|_{q}=(\sum_{i = 1}^{d} |v_{i}|^{q})^{1/q}$ and $|v|=|v|_2$. For any $s > 0$ and a random vector $X$, we say $X\in\L^s$ if $\|X\|_{s} = (\mathbb{E} |X|_{2}^{s})^{1/s} < \infty$. For two positive real or complex sequences $(a_n)$ and $(b_n)$, we say $a_n=\mathcal{O}(b_n)$ or $a_n\lesssim b_n$ (resp. $a_n\asymp b_n$) if there exists $C>0$ such that $|a_n|/|b_n|\le C$ (resp. $1/C\le |a_n|/|b_n|\le C$) for all large $n$, and write $a_n=o(b_n)$ or $a_n\ll b_n$ if $|a_n|/|b_n|\rightarrow0$ as $n\rightarrow\infty$. 

\section{SGD Quantile Estimators as a Markov Chain}\label{setup}
\begin{figure}[!htb]
    \centering
	\includegraphics[width=1\linewidth]{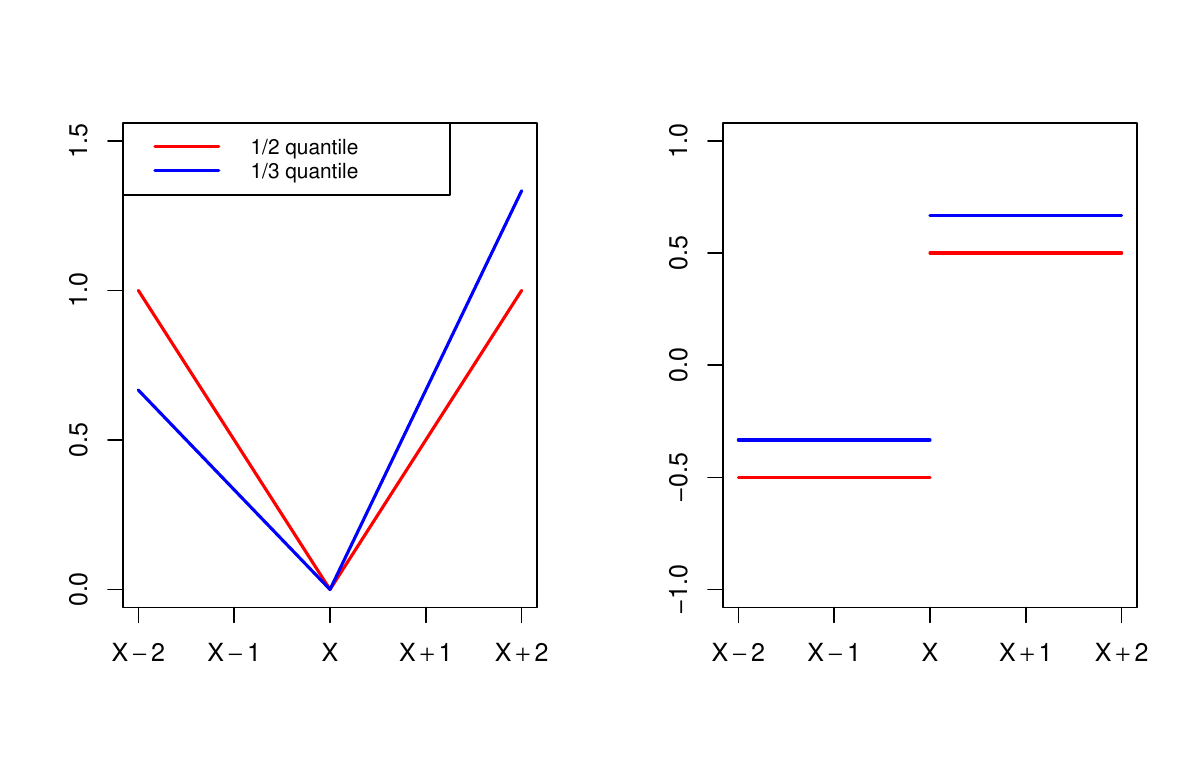}
	\caption{Quantile loss function (left panel) and score function (right panel).}\label{loss}
\end{figure}

Recall the SGD iterations of quantile estimation in \eqref{sgd}. The noise-perturbed loss function $(X-\theta)(\tau-\mathbbm{1}_{\theta \ge X})$ with the sub-gradient $\mathbbm{1}_{\theta \ge X}-\tau$ is neither smooth nor strongly convex, which poses challenges for investigating the limiting distribution of the SGD iterates $\theta_{n+1}(\tau)$. In Figure~\ref{loss}, we provide the quantile loss function and the score function for $1/2$ and $1/3$ quantiles, respectively. 

Moreover, in the previous literature on the asymptotics of non-convex SGD, a dissipativity assumption is usually imposed as a relaxation of strong convexity; see, for example, Assumption 2 in \cite{yu_analysis_2021}. However, quantile loss does not satisfy this condition, and therefore, new theoretical tools are in demand for this particular type of SGD to provide asymptotic properties.

To this end, we interpret the SGD recursion~\eqref{sgd} as a time-homogeneous Markov chain and propose novel techniques adapted from the characteristic functions. Specifically, we consider rational quantile levels $\tau =p/q $ where $p \in \mathbb{N}^+$ and $q \in \mathbb{N}^+$ are mutually prime integers. All possible states of this Markov chain are contained in the set 
$$ \mathcal{M}(\tau)=\Big\{ \theta_{0}(\tau)+\frac{k\eta}{q}\Big\}_{k \in \mathbb{Z}}, $$
where $\theta_{0}(\tau)$ is the initial point. In this paper, we are interested in stationary solutions and the distributional convergence of the SGD iterates $\theta_{n}(\tau)$ as $n\rightarrow\infty$, and the CLT of the stationary distribution as the learning rate $\eta\rightarrow0$. For simplicity, we let $x_0 = \arg\min_{x \in  \mathcal{M}(\tau)}|x-\theta(\tau)|$, and define $x_k=x_0+k\eta/q$ to be the $k$-th state of the Markov chain. In other words, $x_0$ is the state closest to the true quantile, and we would expect the SGD iterate to converge to some distribution centered near $x_0$.

Let $F_{k}=F(x_0+k\eta/q)$ denote the cumulative distribution at the $k$-th state. It is clear that the transition probability from state $x_{s'}$ to $x_s$ of the Markov chain defined in equation \eqref{sgd}, denoted by $P_{s',s}$, satisfies
$$P_{s',s}=
\begin{cases} 
	F_{s'}, & \text{if } s-s' = p-q, \\
	1-F_{s'}, & \text{if }  s-s' = p, \\
	0, & \text{otherwise}.
\end{cases}
$$
To provide the intuition of our proposed methodology, we first suppose that the stationary distribution exists, which will later be shown in Proposition~\ref{markovchain}. Denote the stationary probability of state $x_s$ as $\pi_s$. By definition, it satisfies the following equation 
\begin{equation}\label{station}
    \pi_s=\pi_{s+q-p}F_{s+q-p}+ \pi_{s-p}(1-F_{s-p}), \ \ s\in\mathbb{Z}.
\end{equation}
A concise example is the median estimation, i.e.,  $\tau=1/2$.  In this case, the Markov chain simply moves $\eta/2$ forward when the new sample is greater than the current iterate, or $\eta/2$ backward otherwise. The transition probability matrix is

\[
\begin{bmatrix}
	\ddots & \vdots & \vdots & \vdots &\vdots &\vdots& \iddots \\
	\cdots & 0 & 1-F_{-2} & 0& 0 & 0& \cdots \\
	\cdots & F_{-1} & 0 & 1-F_{-1} &0 & 0&\cdots \\
	\cdots & 0& F_{0}& 0&1-F_{0} &0 &\cdots \\
	\cdots & 0 & 0 & F_{1} & 0&1-F_{1} &\cdots \\
	\cdots & 0& 0& 0 &F_{2} &0&\cdots \\
	\iddots& \vdots & \vdots & \vdots  &\vdots&\vdots& \ddots
\end{bmatrix}.
\]
The Markov chain is almost identical to the birth-and-death process except that it does not have an absorbing state. In this case, equation \eqref{station} becomes
$$\pi_s=\pi_{s+1}F_{s+1}+\pi_{s-1}(1-F_{s-1}),$$
which can be rewritten as
\begin{align}\label{stationary2}
	\pi_s (1-F_{s})-\pi_{s+1}F_{s+1}=\pi_{s-1}(1-F_{s-1})-\pi_{s}F_{s}.
\end{align}
Since $\sum_{s=-\infty}^{\infty} \pi_s=1$ and $F_{s} \leq 1$, both sides of equation \eqref{stationary2} must be $0$, and we have $	\pi_s (1-F_{s})=\pi_{s+1}F_{s+1}.$ In other words, the Markov chain of online median estimation is reversible. This equation has a closed-form solution:
\begin{align*}
\pi_0&=\frac{1}{1+\sum_{i=1}^{\infty}\prod_{j=0}^{i-1}\rho_j+\sum_{i=-1}^{-\infty}\prod_{j=i}^{-1} \rho_j^{-1}},\\
\pi_s&=\pi_0\prod_{j=0}^{s-1}\rho_j, \ s>0 \ \text{and} \ \pi_s=\pi_0\prod_{j=s}^{-1}\frac{1}{\rho_j}, \ s<0,
\end{align*}
where $\rho_j=(1-F_{j})/F_{j+1}$. However, it is still not clear how the stationary distribution evolves when the learning rate $\eta \rightarrow 0$. Moreover, for any other $\tau \neq 1/2$, we do not have such a closed-form stationary probability distribution due to the lack of reversibility, which makes the problem more complicated. Figure \ref{fig:markov_chain} shows the transition probability of the Markov chain with $\tau=1/3$.

\begin{figure}[!t]
    \centering
    \resizebox{\columnwidth}{!}{% 
    \begin{tikzpicture}[line cap=round,line join=round,>=triangle 45,x=1.0cm,y=1.0cm]
        \draw  (-4,4) rectangle (10,-2);
        \clip(-4,-1.3) rectangle (10,3.58);
        
        \node (no) [draw, circle, minimum size=1cm] at (-3,1cm) {$x_{-3}$};
        \node (zero) [draw, circle, minimum size=1cm] at (-1,1cm) {$x_{-2}$};
        \node (one) [draw, circle, minimum size=1cm] at (1,1cm) {$x_{-1}$};
        \node (two) [draw, circle, minimum size=1cm] at (3,1cm) {$x_0$};
        \node (enminusone) [draw, circle, minimum size=1cm] at (5,1cm) {$x_1$};
        \node (en) [draw, circle, minimum size=1cm] at (7,1cm) {$x_2$};
        \node (enplusone) [draw, circle, minimum size=1cm] at (9,1cm) {$x_3$};
        
        \draw [->] (no) .. controls +(0.5,1.5) and +(-0.5,1.5) .. node [midway, above] {$1-F_{-3}$} (zero);
        \draw [->] (zero) .. controls +(0.5,1.5) and +(-0.5,1.5) .. node [midway, above] {$1-F_{-2}$} (one);
        \draw [->] (one) .. controls +(0.5,1.5) and +(-0.5,1.5) .. node [midway, above] {$1-F_{-1}$} (two);
        \draw [->] (two) .. controls +(0.5,1.5) and +(-0.5,1.5) .. node[midway, above]{$1-F_{0}$}(enminusone);
        \draw [->] (enminusone) .. controls +(0.5,1.5) and +(-0.5,1.5) .. node [midway, above] {$1-F_{1}$} (en);
        \draw [->] (en) .. controls +(0.5,1.5) and +(-0.5,1.5) .. node [midway, above] {$1-F_{2}$} (enplusone);
        
        \draw [->] (enplusone) .. controls +(-0.5,-1.5) and +(0.5,-1.5) .. node [midway, below] {$F_{3}$} (enminusone);
        \draw [->] (en) .. controls +(-0.5,-1.5) and +(0.5,-1.5) .. node[midway, below]{$F_{2}$}(two);
        \draw [->] (enminusone) .. controls +(-0.5,-1.5) and +(0.5,-1.5) .. node [midway, below] {$F_{1}$} (one);
        \draw [->] (two) .. controls +(-0.5,-1.5) and +(0.5,-1.5) .. node [midway, below] {$F_{0}$} (zero);
        \draw [->] (one) .. controls +(-0.5,-1.5) and +(0.5,-1.5) .. node [midway, below] {$F_{-1}$} (no);
    \end{tikzpicture}%
    }
    \caption{State transition diagram of the Markov chain.}
    \label{fig:markov_chain}
\end{figure}

Before presenting our first main result, we begin with some basic properties of Markov chain for general quantiles. The Markov chain induced by quantile SGD with $\tau=p/q$ has period $q$ since it can only return to the initial state after $q$ steps. It is also irreducible in the following sense: Let $k_{+}$ and $k_{-}$ denote the maximal and minimal index of the state with the cumulative distribution strictly smaller than $1$ and greater than $0$, i.e.,
$$k_{+}=\max \{k \in \mathbb{Z} :F_{k}<1\}, \ \ k_{-}=\min \{k \in \mathbb{Z} :F_{k}>0\}. $$
Here $k_{+}$ and $k_{-}$ can be $\infty$ and $-\infty$. Since $p$ and $p-q$ are coprime, integer solutions $(m_1,m_2)$ to the linear Diophantine equation $m_1p+m_2(q-p)=k$ always exist for any $k \in \mathbb{N}^+$, which means that there exist paths connecting every two states in this Markov chain. Moreover, the monotonicity of $F$ ensures that the state pair $(x_{k_1},x_{k_2})$ is accessible to each other if and only if $k_{-}+p-q \leq k_1,k_2\leq k_{+}+p$.

 An irreducible and positive recurrent Markov chain has a unique stationary distribution; see, e.g., Theorem 21.13 in \cite{levin2017markov}. We leverage Foster's lemma to prove that the Markov chain \eqref{sgd} is positive recurrent. Once it is done, the periodic convergence in Proposition \ref{markovchain} is an immediate consequence. For convenience, we state Foster's lemma below.

\begin{lemma}[Foster's Lemma]\label{foster}
For an irreducible Markov chain $\{Z_n\}_{n \in \N}$ on a countable state space $\Theta$, suppose that there exists a function  \( L : \Theta \to \mathbb{R}^+ \) such that for some finite set $\mathcal{F}$ and \( \epsilon > 0 \),
\[
\mathbb{E}[L(Z_n) \mid Z_{n-1}] < \infty, \quad \text{for all } Z_{n-1} \in \mathcal{F} ,
\]
\[
\mathbb{E}[L(Z_n) - L(Z_{n-1}) \mid Z_{n-1}] < -\epsilon, \quad \text{for all } Z_{n-1} \not\in  \mathcal{F},
\]
then $\{Z_n\}_{n \in \N}$ is positive recurrent.
\end{lemma}

% Lemma \ref{foster} can be referred to in \cite{Bremaud1999}. 

% {\color{blue}A multivariate version of Foster's Lemma is provided in Section~\ref{sec_joint_CLT}, where extend the univariate $\theta(\tau)$ in~\eqref{eq_minimizer} to a multi-dimensional case.}
The Markov chain $\{\theta_{n}(\eta)\}_{n \in \N}$ is irreducible since $p$ and $q$ are mutually prime. Then it suffices to prove that $\{\theta_{n}(\eta)\}_{n \in \N}$ is also positive recurrent. To this end, we apply Lemma \ref{foster} to verify stability conditions for Markov chains. In particular, a Lyapunov function $L(\theta)$ will be constructed to quantify the chain's deviation from stability. The key idea is to show that, for sufficiently large states, the expected drift of $L(\theta)$ decreases by a fixed amount, ensuring that the chain tends to move back toward smaller, stable states over time. Additionally, it can be shown that the set of states where $L(\theta)$ is small is finite, and the function is bounded in expectation at initialization. These properties collectively satisfy Foster's conditions, proving that the Markov chain returns to a stable region infinitely often and remains well-behaved in the long term. As such, we expect to achieve the following proposition, which demonstrates that the Markov chain of constant learning-rate SGD defined in \eqref{sgd} is positive-recurrent with no further assumptions.

\begin{proposition}[Stationary distribution]\label{markovchain}
	Consider the quantile estimates $\{\theta_{n}(\tau)\}_{n \in \N}$ in~\eqref{sgd}. The recursion forms a Markov chain with a unique stationary distribution $\mathcal{P}_{\eta}$. Moreover, let $\mathcal{P}_{\eta}(y)$ denote the stationary probability of some state $y$. For any initial point $\theta_{0}(\tau)$, let $S_0,S_1,\ldots,S_{q-1}$ be the cyclic decomposition of the state space, where
	$$ S_j = \{\theta_{0}(\tau)+k\eta: k \in \mathbb{Z}, \, k \equiv jp \pmod q \},$$
	for $j=0,1,\ldots,q-1$. Then, for all $y \in S_j$, as $n\rightarrow\infty$,
    $$ \mathbb{P}(\theta_{nq+j}(\tau)=y) \rightarrow q\mathcal{P}_{\eta}(y).$$
\end{proposition}

\begin{remark}[Initialization]\label{init}
With a fixed initial point, the periodic Markov chain does not converge to its stationary distribution because its support varies from the $n$-th step to the $(n+1)$-th step. However, in practice we can randomize the choice of initial points as a uniform distribution over $\{ \theta_{0}, \theta_{0}+\eta/q,  \theta_{0}+2\eta/q,\ldots,  \theta_{0}+(q-1)\eta/q\}$. Then following Proposition \ref{markovchain}, the SGD sequence \eqref{sgd} weakly converges to the stationary distribution $\mathcal{P}_{\eta}$.
\end{remark}

\begin{conjecture}[Irrational quantile levels]\label{conjecture_irrational}
For rational quantile levels, Proposition \ref{markovchain} resolves the problem of the existence of stationarity and the weak convergence for SGD quantile estimates with constant learning rates. The convergence problem for SGD quantile estimates with irrational levels is very different and mathematically more challenging. Consider the quantile SGD with irrational level $\tau \in (0,1)$ and learning rate $\eta$. For simplicity, let $\theta_0 (\tau) = 0$ and assume that the distribution of $X$ is supported on $\R$. The Markov chain either moves $\eta \tau$ or $\eta(\tau-1)$ every time, and the state space becomes
$$ \mathcal{M}(\tau)=\Big\{ \eta(n\tau-m): \ \ n,m \in \mathbb{N}, \,\ m \leq n \Big\}.$$
This state space is still countable but dense in $\mathbb{R}$. Consequently, the results for discrete state space Markov chains do not apply here. Another key observation is that the Markov chain with irrational $\tau$ never returns to any state it has visited. As a result, there is no stationary distribution defined on $\mathcal{M}(\tau)$. So we need to study this scenario as a continuous state Markov chain, and investigate the stationary measure defined on $\R$.

The Markov chain is not irreducible on $\R$ since it only has countable accessible points, so we can not use any result based on the irreducibility. However, since we have shown that the stationary measure exists for any rational level, we can take a rational sequence $p_l/q_l \rightarrow \tau$ as $l \to \infty$ with the corresponding $F_{\text{stationary},l}(x)$, the cumulative distribution function of the stationary measure with the rational quantile level $p_l/q_l$. We conjecture that this $F_{\text{stationary},l}(x)$ converges to some cumulative distribution function $F(x)$ (say), and the limiting function is the distribution of the stationary measure with the irrational quantile level $\tau$.
\end{conjecture}

\section{Theoretical Results}\label{Theory}
In this section, we investigate the asymptotic performance of the stationary distribution. We first centralize and standardize the Markov chain. In particular, we consider $\tilde{x}_k=(x_k-x_0)/\sqrt{\eta}$ as the new $k$-th state. Here and in the sequel, $\{\tilde{x}_k\}_{k \in \mathbb{Z}}$ will represent the new standardized state space, i.e., 
$$\tilde{x}_k=\frac{k\sqrt{\eta}}{q}, k \in \mathbb{Z},$$
$\pi_{\eta,k}$ represents the stationary probability of the standardized Markov chain at the $k$-th state, and $\mathcal{P}_{\eta}$ denotes the stationary distribution of the centered and standardized Markov chain. To show that $\mathcal{P}_{\eta}$ is asymptotically normal when $\eta \rightarrow 0$, we first assume a regularity condition on the density of $X$, which is standard in the quantile literature. 

\begin{assumption}[Density]\label{asm_density}
	Recall $\theta(\tau)$ defined in~\eqref{eq_minimizer}. Assume that the random variable $X$ has a density function $f_X$ being $C^2$ smooth in an interval $\mathcal{B}_r(\theta(\tau))= [\theta(\tau)-r, \theta(\tau)+r]$ for some $r>0$, with $f_X(\theta(\tau))>0$.  
\end{assumption}
Assumption~\ref{asm_density} guarantees the existence and uniqueness of the $\tau$-th quantile. We do not impose any requirement for the tail probability or the moment boundedness of the distribution. To prove the CLT result, we first propose the following Lemma \ref{MGFbound} and Corollary \ref{tail} and \ref{Conineq} to bound the tail probability and moments of the stationary distribution.

\begin{lemma}[Moment generating function]\label{MGFbound} Consider the stationary probability $\pi_{\eta,k}$ specified above. Under Assumption~\ref{asm_density}, given any $\beta >3$, for all $\eta$ sufficiently small and $\gamma=0,1,2$, we have 
	$$\sum_{k \in \mathbb{Z}} \eta^{\beta} \pi_{\eta,k} |k|^{\gamma}e^{\frac{|k|\sqrt{\eta}}{q}} \leq q^2.$$
    For $k=0$ and $\gamma=0$, we use the convention that $0^0=1$.
\end{lemma}

Technically, Lemma \ref{MGFbound} provides an upper bound of the moment generating function $\text{MGF}(t)$ of the stationary distribution at $t=1$, as well as its first and second derivatives both at $t=1$. The upper bound has a polynomial rate of $1/\eta$. The following Corollary \ref{tail} and \ref{Conineq} are direct consequences of Lemma \ref{MGFbound}.

\begin{corollary}\label{tail}
	Under Assumption~\ref{asm_density}, given any integer $K_0>\beta$ where $\beta$ is the same as in Lemma \ref{MGFbound}, let $N = 
    \lceil qK_0 \log (1/\eta)/ \sqrt{\eta} \rceil$. Then for all $\eta$ sufficiently small, 
	$$\sum_{|k| \ge N} \pi_{\eta,k} |k|^\gamma \leq q^2 \eta^{K_0-\beta},$$
	where $\gamma=0,1,2$.
\end{corollary}

Notice that when $|k|< N$, $|\tilde{x}_k| \leq K_0\log(1/\eta)$. Corollary \ref{tail} indicates that if we truncate the state space by an $\mathcal{O}(\log(1/\eta))$ rate of the boundary, the moments over the tail region of the stationary distribution decay polynomially fast. The MGF bound also implies the following concentration inequality.

\begin{corollary}[Concentration inequality]\label{Conineq}
	Let $Z$ follow the stationary distribution of the Markov chain induced by $\{ \theta_n(\tau) \}_{n \in \N}$, the original SGD sequence. Under the same conditions in Lemma \ref{MGFbound}, for any $\epsilon >\eta/q $,
    \begin{align*}
    \P (|Z-\theta(\tau)|  \ge \epsilon) &\leq \frac{ \eta^{2-\beta}} {\epsilon^2}\exp\Big( -\frac{\epsilon}{\sqrt{\eta}} +\frac{\sqrt{\eta}}{q}\Big) \\
    &\quad \cdot\frac{q^2}{q^2+\eta^2/\epsilon^2-2q \eta/\epsilon}.
    \end{align*}
\end{corollary}
\begin{remark}
Regarding the statistical properties of recursive quantile estimators, \cite{zhang_piecewise_2025} designed a piecewise Lyapunov function and derived $p$-th moment bounds. In comparison, we bound the MGF and further provide an exponential tail concentration inequality for the stationary distribution of the quantile SGD estimators. Our tail probability bounds do not follow from their results.
\end{remark}
Now we are ready to present the main CLT results. The following Theorem \ref{charac} shows that the characteristic function of $\mathcal{P}_{\eta}$ converges to the characteristic function of Gaussian distributions.

\begin{theorem}[Characteristic function]\label{charac}
	Recall $\theta(\tau)$ defined in~\eqref{eq_minimizer}. Suppose that Assumption~\ref{asm_density} holds. Let $\phi_{\eta}(t)$ denote the characteristic function of the standardized stationary distribution with the learning rate $\eta$. Then we have the following point-wise convergence that for any $t \in \mathbb{R}$,
	$$ \lim_{\eta \rightarrow 0} \phi_{\eta}(t)=e^{-\tfrac{\tau(1-\tau)t^2}{4f_X(\theta(\tau))}}.$$
\end{theorem}
The asymptotic normality follows directly from Theorem \ref{charac}	and Lévy's continuity theorem.
\begin{proposition}[Asymptotic normality]\label{asympnormal}
	Under the same conditions in Theorem~\ref{charac}, the stationary distribution of $(\theta_{n}(\tau)-x_0)/\sqrt{\eta}$ converges to the following normal distribution,
	$$\mathcal{P}_{\eta} \stackrel{\mathcal{D}}{\to} \mathcal{N}\Big(0, \frac{\tau(1-\tau)}{2f_X(\theta(\tau))}\Big),\quad \text{as }\eta\rightarrow0.$$
\end{proposition}
\begin{remark}[Quenched CLT]
By definition, we have $|x_0-\theta(\tau)|<\eta$. Therefore, Proposition \ref{asympnormal} also implies that the stationary distribution of $(\theta_{n}(\tau)-\theta(\tau))/\sqrt{\eta}$ converges to the same normal distribution regardless of the fixed initial point. In this sense, our result is a quenched version of CLT where the asymptotic distribution does not rely on the initial point.
\end{remark}

\section{Online Inference}

In this section, we propose a recursive kernel density estimator
\begin{equation}\label{kde}
    \hat{f}_n(\tau) = \frac{1}{n} \sum_{k=1}^{n} \frac{K_{b_k}(\theta_{k-1}(\tau), X_k)}{b_k},
\end{equation}
where \( (b_k)_{1\le k\le n} \) is the bandwidth sequence chosen as $b_k \asymp k^{-\alpha}$ for some $0 <\alpha <1$ and \( K_b(x,u) = K((x - u)/b) \) for some kernel function \( K(\cdot) \). We assume that the \( K(\cdot) \)  satisfies the following condition:
\begin{assumption}[Kernel]\label{kernel}
    The kernel \( K \) has a bounded support \([-M,M]\). Assume $\sup_u |K(u)|  < \infty$, $\int_{\mathbb{R}} K(u) du = 1$, $\int_{\mathbb{R}} u^2 |K(u)|du< \infty$, and $\int_{\mathbb{R}} sK(s) ds =0$.
\end{assumption}
Assumption \ref{kernel} is satisfied by many popular choices of kernels such as the rectangle kernel $K(v) = \mathbbm{1}_{|v|<1/2},$ the Epanechnikov kernel $K(v) = 3(1 - v^2) \mathbbm{1}_{|v|<1}/4$ among others. In practice, we can simply take $b_k=k^{-1/5}$ (refer to Theorem 3 in \cite{huang_recursive_2014}). Finally, the quenched CLT in Proposition \ref{asympnormal} similarly holds with $f_X(\theta(\tau))$ therein replaced by the consistent estimator $\hat f_n(\tau)$ by Slutsky's theorem, which is stated as follows.

\begin{theorem}[Consistency of the online kernel estimator]\label{OKDE}
Suppose that conditions in \ref{asm_density} and \ref{kernel} hold, and further assume that the density function $f_X$ is $C^2$ smooth with bounded first and second derivatives on $\R$. Then we have
$$ \lim _{\eta\rightarrow0}\lim _{n \rightarrow \infty} \hat f_n(\tau) = f_X(\theta(\tau)) \quad \text{almost surely.}$$ 
\end{theorem}
As a direct consequence of \ref{OKDE} and Proposition \ref{asympnormal}, we have the following result.
\begin{corollary}[Asymptotic normality for implementation]\label{asympnormal_est}
	Consider the SGD iterates \eqref{sgd} with the initialization choice in Remark \ref{init}. Let $n\rightarrow \infty$ and then $\eta \rightarrow 0$, we have the following weak convergence,
	$$\sqrt{\hat f_n(\tau)}\frac{\theta_{n}(\tau)-\theta(\tau)}{\sqrt{\eta}}\stackrel{\mathcal{D}}{\to} \mathcal{N}\Big(0, \frac{\tau(1-\tau)}{2}\Big).$$
\end{corollary}

\section{Simulation}\label{simulation}

In the simulation study, we estimate the $\tau=3/4$-th quantile of the Beta$(2,3)$ distribution and the Cauchy distribution with scale parameter $2$, using SGD with constant learning rate $\eta=0.01$ and $0.001$. In this way, we validate our results and the online inference method through asymmetric and heavy-tailed distributions. Based on the asymptotic normality result, we construct $100\%(1-\alpha)$ confidence interval of $\theta(\tau)$ as
$$\Bigg[\theta_n(\tau)-z_{1-\alpha/2}\sqrt{\frac{\tau(1-\tau)\eta}{2\hat{f}_n(\tau)}},\theta_n(\tau)+z_{1-\alpha/2}\sqrt{\frac{\tau(1-\tau)\eta}{2\hat{f}_n(\tau)}}\Bigg],$$
where $\hat{f}_n(\tau)$ is the estimated population density at $\theta(\tau)$. We estimate it through the fully online kernel density estimation \eqref{kde}, and assess the performance of empirical coverage with nominal level $\alpha=0.05$.

\begin{figure}[!hbt]
    \centering
    \subfloat[Beta $(2,3)$]{
        \includegraphics[width=0.45\textwidth]{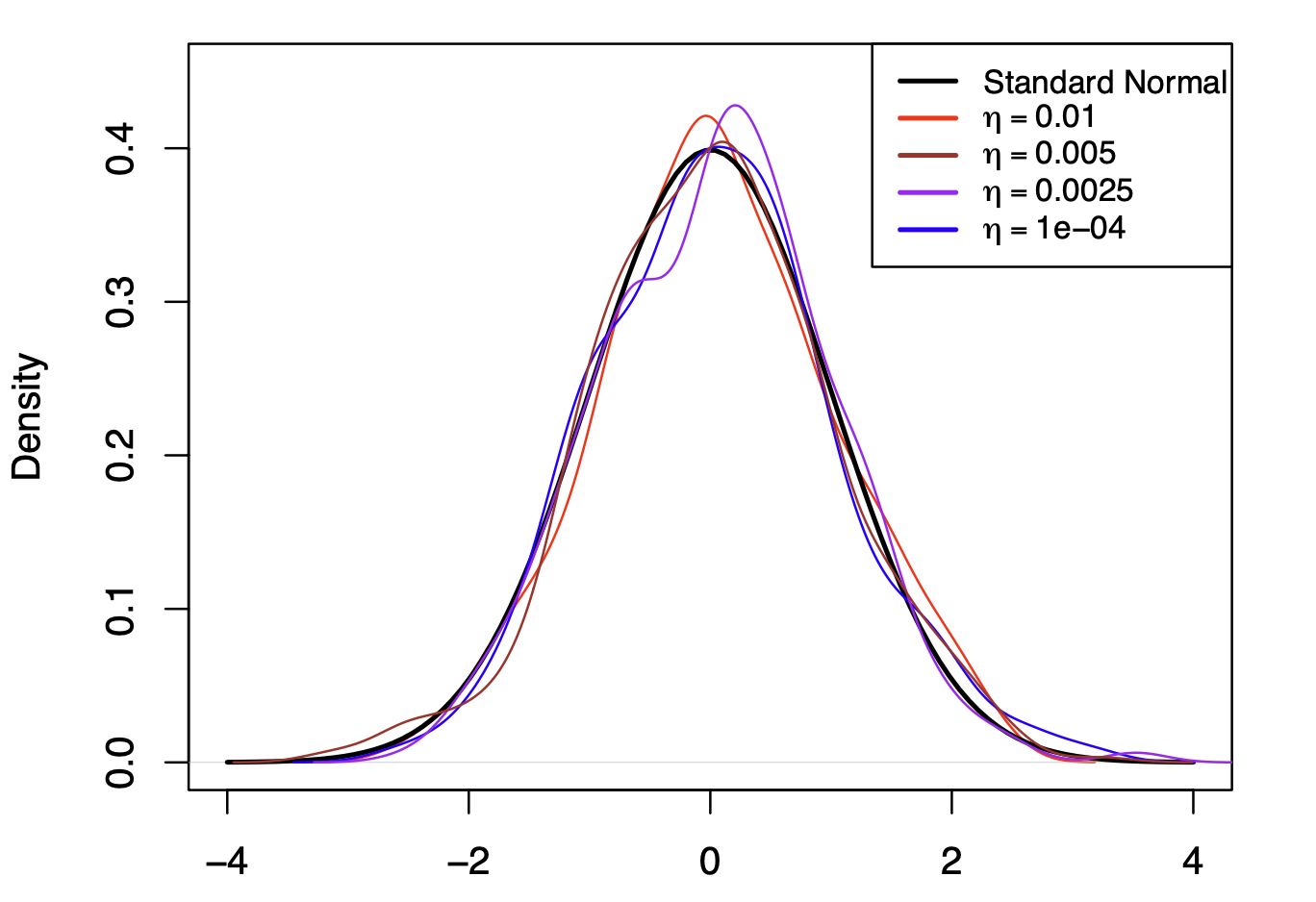}
    }
    \hfil
    \subfloat[Cauchy $(0,2)$]{
        \includegraphics[width=0.45\textwidth]{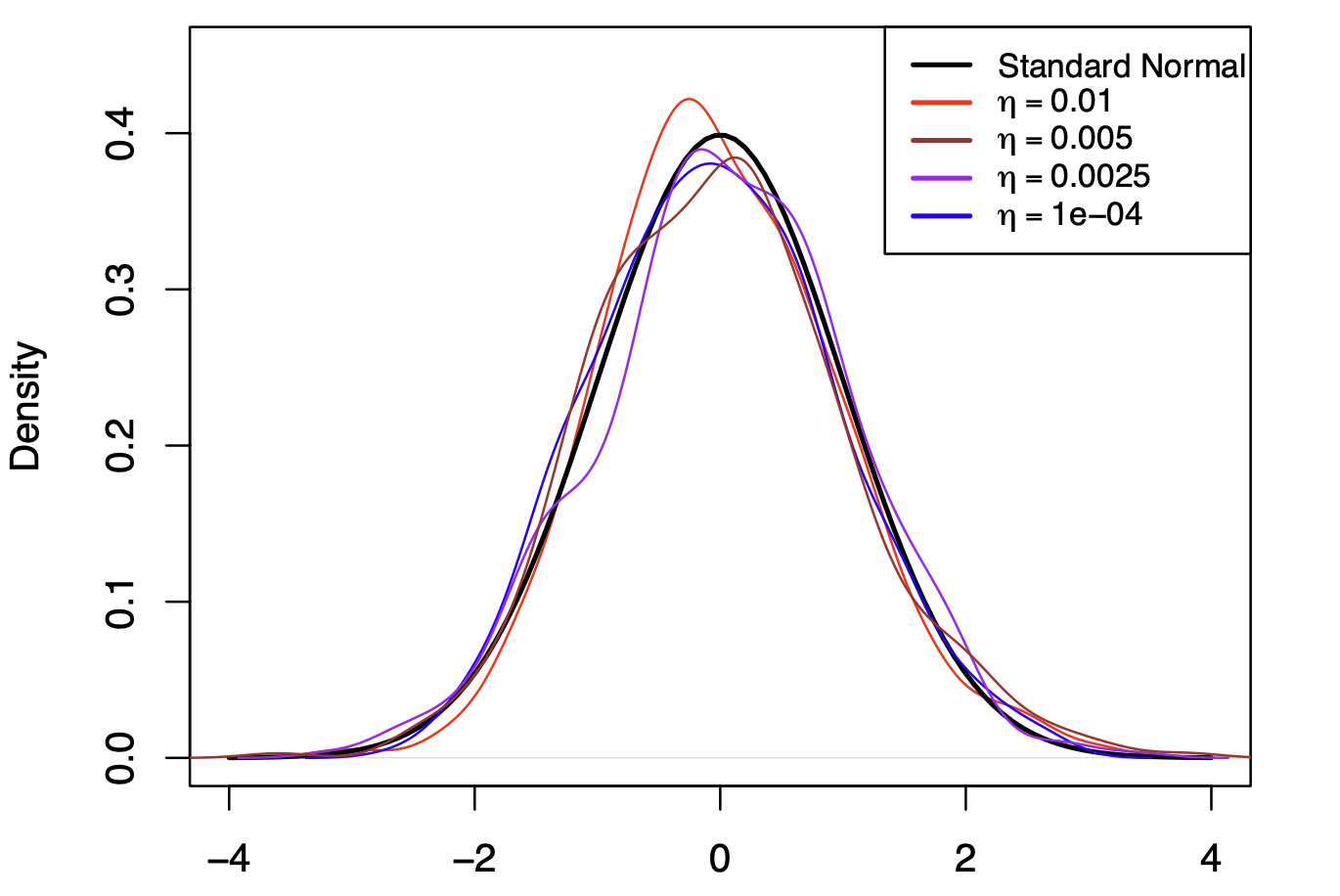}
    }
    \caption{Asymptotic normality for quantiles of Beta $(2,3)$ and Cauchy $(0,2)$ distributions.}
    \label{fig_normal}
\end{figure}

\begin{figure}[!hbt]
    \centering
    \subfloat[Beta $(2,3)$]{
        \includegraphics[width=0.45\textwidth]{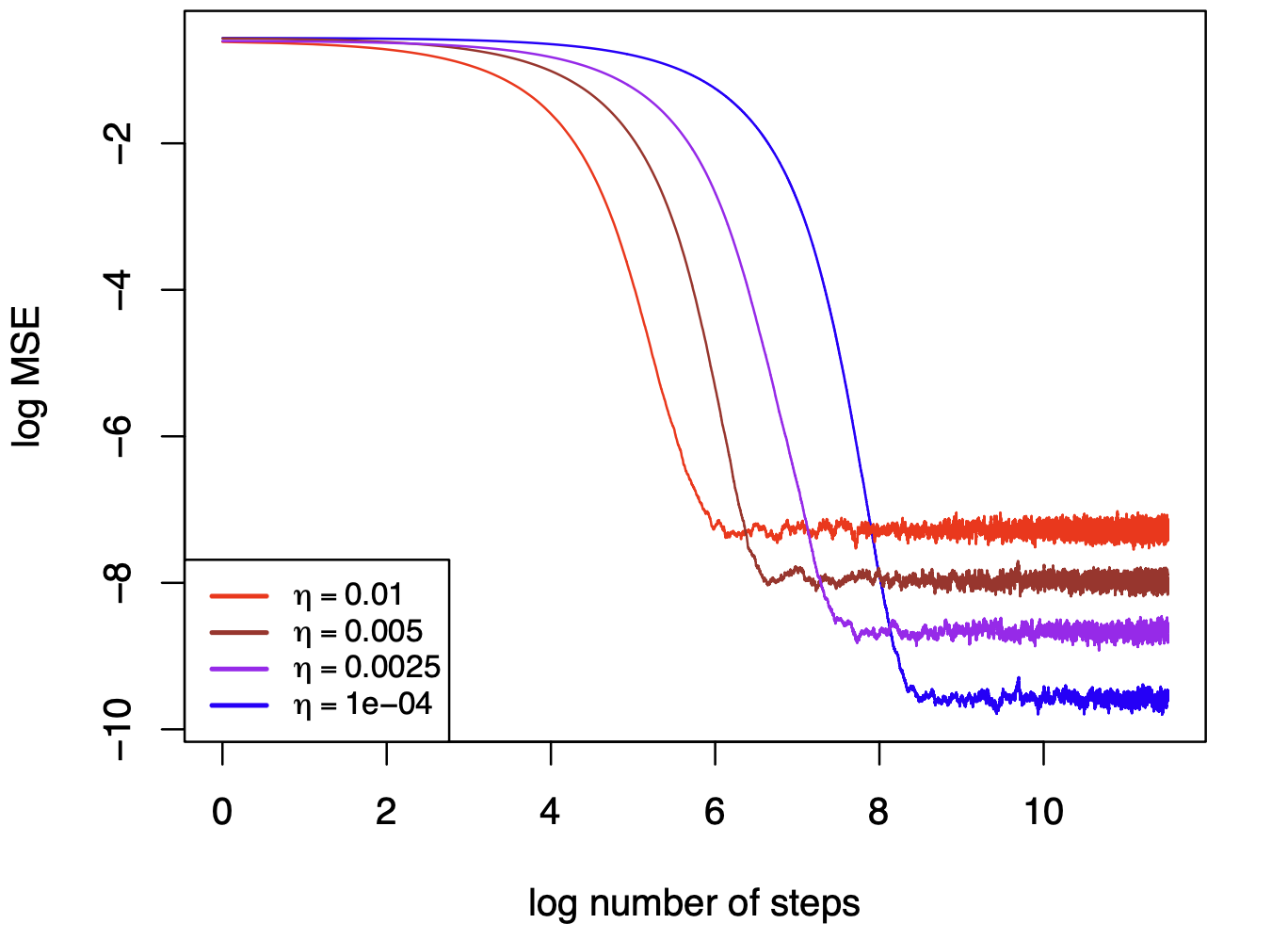}
    }
    \hfil
    \subfloat[Cauchy $(0,2)$]{
        \includegraphics[width=0.45\textwidth]{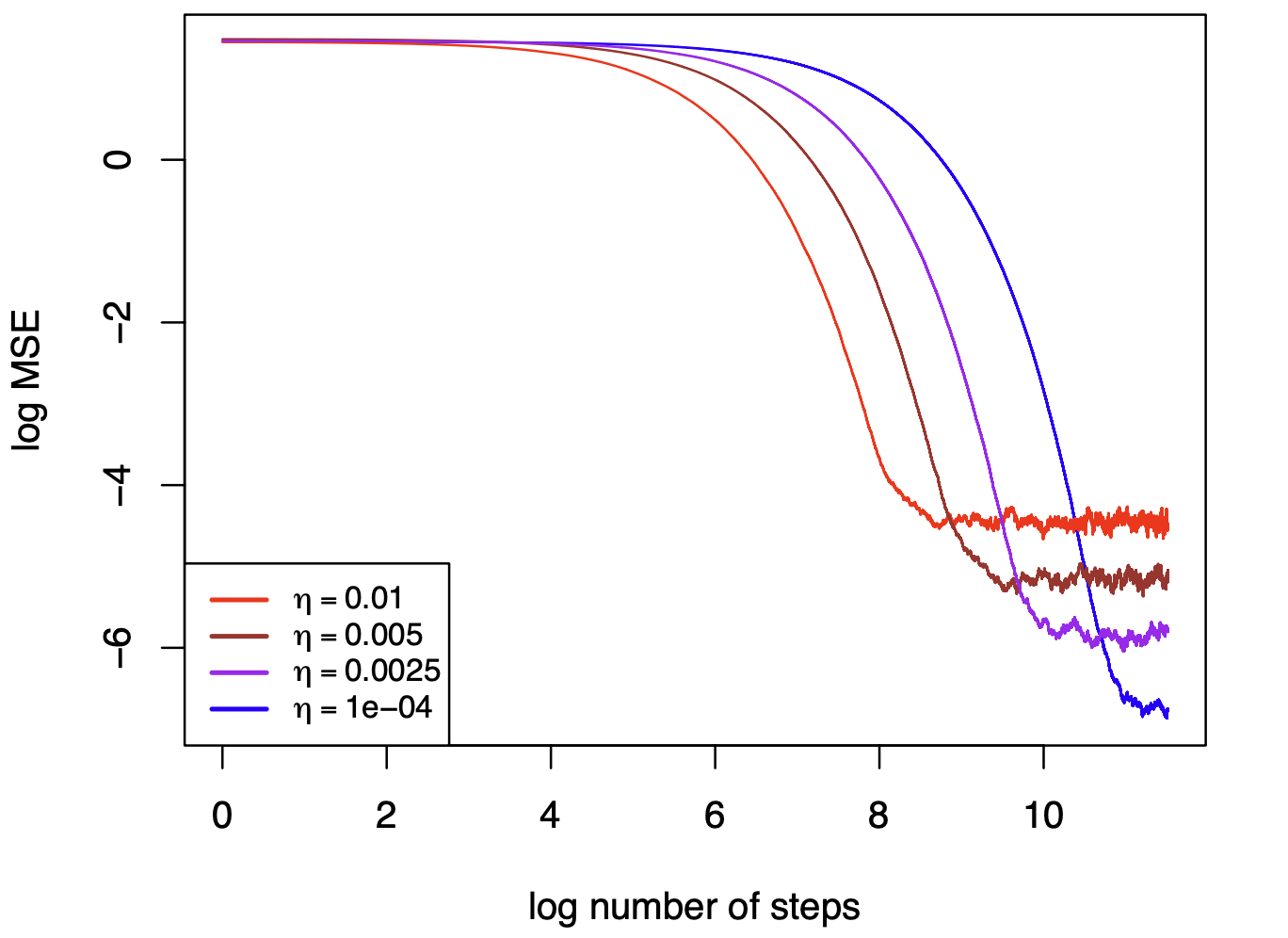}
    }
    \caption{Convergence of mean squared errors for quantiles of Beta $(2,3)$ and Cauchy $(0,2)$ distributions.}
    \label{fig_mse}
\end{figure}

\begin{figure}[!hbt]
    \centering
    \subfloat[Beta $(2,3)$]{
        \includegraphics[width=0.45\textwidth]{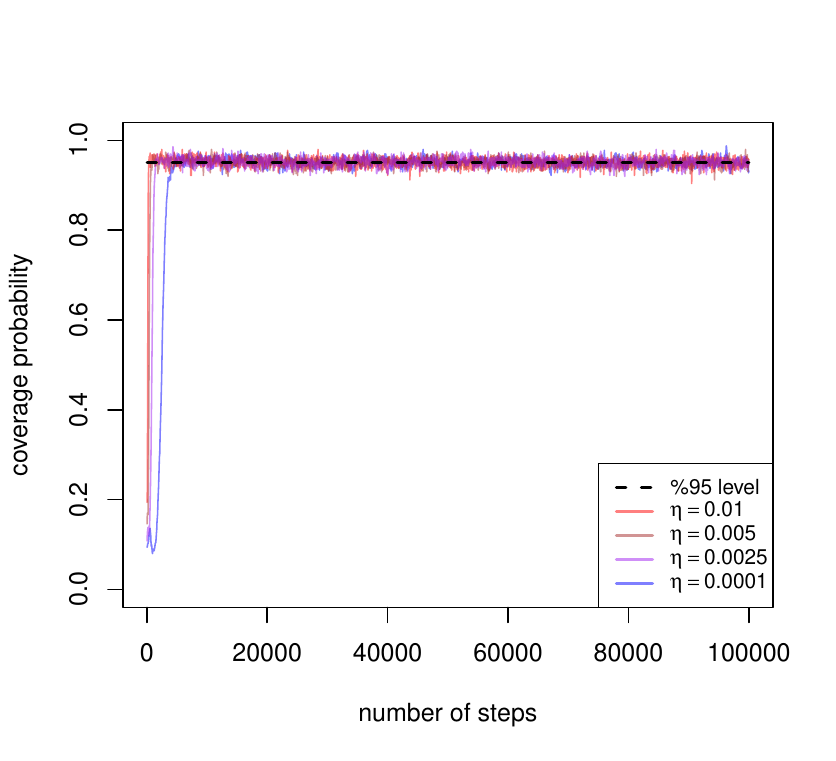}
    }
    \hfil
    \subfloat[Cauchy $(0,2)$]{
        \includegraphics[width=0.45\textwidth]{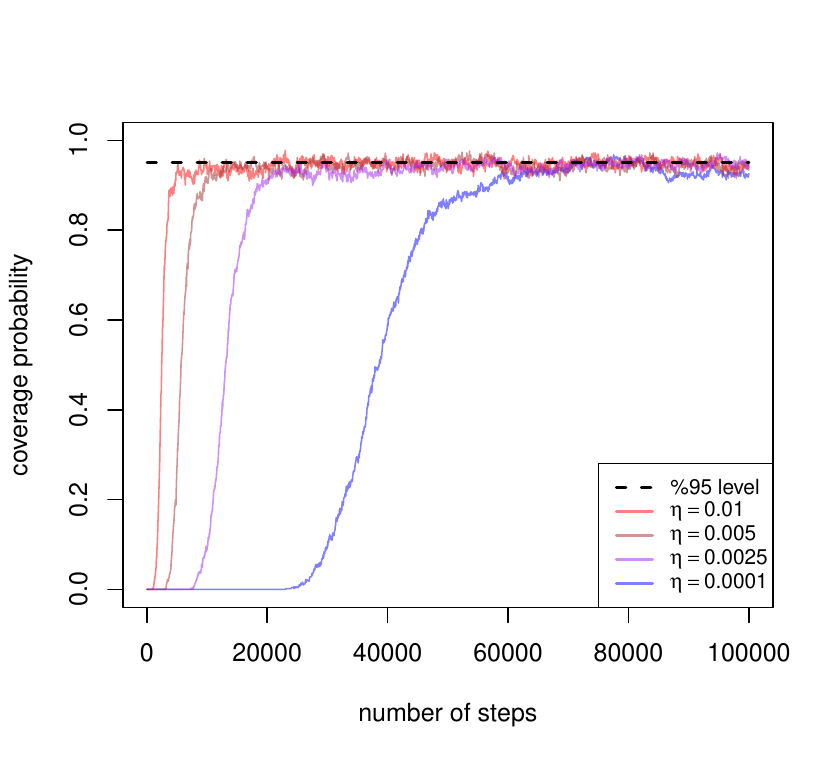}
    }
    \caption{Coverage probabilities for quantiles of Beta $(2,3)$ and Cauchy $(0,2)$ distributions.}
    \label{fig_cov}
\end{figure}

Figures \ref{fig_normal}--\ref{fig_cov} and Table \ref{band} show the asymptotic behavior of the quantile SGD estimator and the performance of the online inference procedure. All results are averaged over 500 independent runs. They support our central limit theorem in Proposition \ref{asympnormal}, confirming that the SGD iterates converge in distribution to a Gaussian with the specified asymptotic variance. Furthermore, the empirical coverage of recursive confidence intervals approaches the nominal level across different learning rates. Overall, these numerical experiments validate both our theoretical findings and the effectiveness of the proposed uncertainty quantification method, even under asymmetric or heavy-tailed data distributions.

\begin{table}[!htp]
	\caption{The Empirical Coverage Probability of the Confidence Intervals by the Online Inference Method.}
	\label{band}
	\centering
    \scriptsize
    \setlength{\tabcolsep}{2pt}
	\begin{tabular}{lcccc}
		\toprule
		& \multicolumn{4}{c}{$\eta=0.01$} \\
		\cmidrule(r){2-5}
		& $n=50000$ & $n=100000$  & $n=150000$  & $n=200000$  \\
		\cmidrule(r){2-5}
		\textbf{Beta Distribution}  &0.954 &0.958 &0.954 &0.932\\
		\textbf{Cauchy Distribution}	&0.940  &	0.938	&0.956 &	0.952 \\
		\cmidrule(r){1-5}
		& \multicolumn{4}{c}{$\eta=0.005$} \\
		\cmidrule(r){2-5}
		& $n=50000$ & $n=100000$  & $n=150000$  & $n=200000$   \\
		\cmidrule(r){2-5}
		\textbf{Beta Distribution} &0.958 &0.946 &0.948 &0.924\\
		\textbf{Cauchy Distribution}	 &0.946 &0.954 &0.954 &0.936\\
        \cmidrule(r){1-5}
		& \multicolumn{4}{c}{$\eta=0.0025$} \\
		\cmidrule(r){2-5}
		& $n=50000$ & $n=100000$  & $n=150000$  & $n=200000$   \\
		\cmidrule(r){2-5}
		\textbf{Beta Distribution}  &0.942 &0.948 &0.944 &0.954\\
		\textbf{Cauchy Distribution}	&0.954 &0.936 &0.936 &0.942\\
        \cmidrule(r){1-5}
		& \multicolumn{4}{c}{$\eta=0.001$} \\
		\cmidrule(r){2-5}
		& $n=50000$ & $n=100000$  & $n=150000$  & $n=200000$   \\
		\cmidrule(r){2-5}
		\textbf{Beta Distribution}   &0.962 &0.932 &0.944 &0.956\\
		\textbf{Cauchy Distribution}	&0.858    &	0.926 	&0.946&	0.960 \\
		\bottomrule
	\end{tabular}
\end{table}

\section{Proof Sketch of Lemma \ref{MGFbound}}
This section outlines the main techniques we used in the proof of Lemma \ref{MGFbound}. We do the following three steps.

    \textbf{Step 1.} Motivated by Theorem 1 in \cite{tweedie1983existence}, we first prove the following Lemma. 
    \begin{lemma}\label{tweedie}
    Let $\{Z_n\}_{n \in \N}$ be a positive recurrent Markov chain with countable state space $\mathscr{X} = \{x_k\}_{k \in \mathbb{Z}}$ and stationary distribution $\{\pi_k\}_{k \in \mathbb{Z}}$. Given a set $\mathcal{A} \subseteq \mathscr{X} $ with positive stationary probability and some non-negative measurable functions $g$ and $f$, suppose that for any $z\in \mathcal{A}^c$, we have
    \begin{equation}\label{condition}
        \max\{\E [g(Z_1) \mathbbm{1}_{ Z_1 \in \mathcal{A}^c}\mid Z_0=z],0\} \leq g(z)-f(z), 
    \end{equation} 
    then
    $$ \E_{\pi} [f(Z) \mathbbm{1}_{ Z \in \mathcal{A}^c}] \leq \sup_{z \in \mathcal{A}} \E[g(Z_1) \mathbbm{1}_{ Z_1 \in \mathcal{A}^c}\mid Z_0=z], $$
    where $\E_{\pi}$ denotes the expectation under the stationary distribution of $\{Z_n\}_{n \in \N}$.
    \end{lemma}
    The main application of Lemma \ref{tweedie} is to control the stationary expectation of some functional $f$ of a positive recurrent Markov chain by its dominant function $g$. Usually, $\mathcal{A}$ is chosen as a finite or tractable set, and the conclusion of Lemma \ref{tweedie} can be used to bound the expectation over $\mathcal{A}^c$ by the function value over $\mathcal{A}$.
    \begin{proof}
    Define 
    $T_{\mathcal{A}}=\inf\{n \ge 1: Z_n \in \mathcal{A} \}$ as the hitting time on $\mathcal{A}$. Notice that for $n\ge 2 $ and $x_k \in \mathcal{A}^c$ we have
\begin{align*} & \qquad\P (Z_n=x_k,T_\mathcal{A} \ge n \mid Z_0=x_j)\\
&=\sum_{l: x_l\in \mathcal{A}^c} \P(Z_{n-1}=x_l,T_\mathcal{A} \ge n-1 \mid Z_0=x_j) \\
&\qquad\qquad\qquad \cdot \P(Z_1=x_k \mid Z_0=x_l).
\end{align*}
Now we consider the following inequality:
\begin{align*}
  0 &\leq \sum_{k: x_k \in \mathcal{A}^c} \P(Z_n=x_k, T_\mathcal{A} \ge n \mid Z_0=x_j) g(x_k)\\
  &= \sum_{k: x_k \in \mathcal{A}^c} \Big(\sum_{l: x_l\in \mathcal{A}^c}
      \P(Z_{n-1}=x_l, T_\mathcal{A} \ge n{-}1 \mid Z_0=x_j) \\
  &\qquad\qquad\qquad \cdot \P(Z_1=x_k \mid Z_0=x_l)\Big) g(x_k)\\
  &= \sum_{l: x_l\in \mathcal{A}^c} \P(Z_{n-1}=x_l, T_\mathcal{A} \ge n{-}1 \mid Z_0=x_j) \\
  &\qquad \cdot \Big(\sum_{k: x_k \in \mathcal{A}^c} \P(Z_1=x_k \mid Z_0=x_l) g(x_k)\Big)\\
  &= \sum_{l: x_l\in \mathcal{A}^c} \P(Z_{n-1}=x_l, T_\mathcal{A} \ge n{-}1 \mid Z_0=x_j) \\
  &\qquad \cdot \E[g(Z_1)\mathbbm{1}_{Z_1 \in \mathcal{A}^c} \mid Z_0=x_l]\\
  &\leq \sum_{l: x_l\in \mathcal{A}^c} \P(Z_{n-1}=x_l, T_\mathcal{A} \ge n{-}1 \mid Z_0=x_j) \\
  &\qquad \cdot (g(x_l) - f(x_l)).
\end{align*}
We iteratively use the inequality and obtain
\begin{align*}
& \quad \sum_{k: x_k\in \mathcal{A}^c}\P(Z_{n-1}=x_k,T_\mathcal{A} \ge n-1 \mid Z_0=x_j)\\
&\qquad\qquad\qquad \cdot (g(x_k)-f(x_k))\\
&\leq \sum_{k: x_k\in \mathcal{A}^c}\P(Z_{n-2}=x_k,T_\mathcal{A} \ge n-2\mid Z_0=x_j)g(x_k)\\
&-\sum_{m=n-2}^{n-1}\Big(\sum_{k: x_k\in \mathcal{A}^c}\P(Z_m=x_k,T_\mathcal{A} \ge m\mid Z_0=x_j)f(x_k)\Big)\\
&\leq \ldots\\
&\leq \sum_{k:  x_k\in \mathcal{A}^c}\P(Z_{1}=x_k,T_\mathcal{A} \ge 1\mid Z_0=x_j)g(x_k)\\
& \;\,-\sum_{m=1}^{n-1}\Big(\sum_{k:  x_k\in \mathcal{A}^c}\P(Z_m=x_k,T_\mathcal{A} \ge m\mid Z_0=x_j)f(x_k)\Big)\\
&= \sum_{k: x_k\in \mathcal{A}^c}\P(Z_{1}=x_k\mid Z_0=x_j)g(x_k)\\
&\;\,-\sum_{m=1}^{n-1}\Big(\sum_{k: x_k\in \mathcal{A}^c}\P(Z_m=x_k,T_\mathcal{A} \ge m\mid Z_0=x_j)f(x_k)\Big)\\
&= \E[g(Z_1)\mathbbm{1}_{Z_1 \in \mathcal{A}^c}\mid Z_0=x_j]\\
&\;\,-\sum_{m=1}^{n-1}\Big(\sum_{k: x_k\in \mathcal{A}^c}\P(Z_m=x_k,T_\mathcal{A} \ge m\mid Z_0=x_j)f(x_k)\Big)%\\
%&\leq g(x_j)-f(x_j)\\
%& \;\,-\sum_{m=1}^{n-1}\Big(\sum_{k: x_k\in \mathcal{A}^c}\P(Z_m=x_k,T_\mathcal{A} \ge m\mid Z_0=x_j)f(x_k)\Big).
\end{align*}
Since
\begin{align*}
    &\sum_{k: x_k\in \mathcal{A}^c}\P(Z_{n-1}=x_k,T_\mathcal{A} \ge n-1 \mid Z_0=x_j)\\
&\qquad\qquad\qquad \cdot (g(x_k)-f(x_k)) \ge 0,
\end{align*}
we have
\begin{align*}
&\sum_{m=1}^{n-1}\Big(\sum_{k: x_k\in \mathcal{A}^c}\P(Z_m=x_k,T_\mathcal{A} \ge m\mid Z_0=x_j)f(x_k)\Big) \\
&\qquad \leq \E[g(Z_1)\mathbbm{1}_{Z_1 \in \mathcal{A}^c}\mid Z_0=x_j].
\end{align*}
Let $n\rightarrow\infty$, 
\begin{align*}
&\sum_{k: x_k\in \mathcal{A}^c}f(x_k) 
\Big(\sum_{m=1}^{\infty}\P(Z_m=x_k,T_\mathcal{A} \ge m\mid Z_0=x_j)\Big) \\
&\qquad \leq \E[g(Z_1)\mathbbm{1}_{Z_1 \in \mathcal{A}^c}\mid Z_0=x_j].
\end{align*}
For $k$ such that $x_k \notin \mathcal{A}$, the stationary distribution $\pi_k$ has the following representation (\cite{tweedie1983existence}),
\begin{equation}\label{repre}
    \pi_k=\sum_{j:  x_j \in \mathcal{A}} \pi_j \Big(\sum_{n=1}^{\infty}\P (Z_n=x_k,T_\mathcal{A} \ge n \mid Z_0=x_j)\Big).
\end{equation}
Finally, we plug equation \eqref{repre} into the expression of $\E_{\pi} [f(Z) \mathbbm{1}_{ Z \in \mathcal{A}^c}]$, and get
\begin{align*} 
&\quad \E_{\pi} [f(Z) \mathbbm{1}_{ Z \in \mathcal{A}^c}] \\
&= \sum_{k: x_k \in \mathcal{A}^c}f(x_k)\pi_k\\
&= \sum_{k: x_k \in \mathcal{A}^c}f(x_k) \\
 &\qquad \cdot \Big[\sum_{j: x_j \in \mathcal{A}} \pi_j \Big(\sum_{n=1}^{\infty}\P(Z_n=x_k,T_\mathcal{A} \ge n \mid Z_0=x_j)\Big)\Big]\\
&= \sum_{j: x_j \in \mathcal{A}} \pi_j\\
 &\qquad\cdot\Big[\sum_{k: x_k \in \mathcal{A}^c}f(x_k)
\Big(\sum_{n=1}^{\infty}\P(Z_n=x_k,T_\mathcal{A} \ge n \mid Z_0=x_j)\Big)\Big]\\
&\leq \sum_{j: x_j \in \mathcal{A}} \pi_j\Big[\E[g(Z_1)\mathbbm{1}_{Z_1 \in \mathcal{A}^c}\mid Z_0=x_j]\Big] \\
&\leq \sup_{z \in \mathcal{A}} \E[g(Z_1)\mathbbm{1}_{Z_1 \in \mathcal{A}^c}\mid Z_0=z],
\end{align*}
which is the conclusion we aim to prove.
\end{proof}
\textbf{Step 2.} \label{Lstep2}We only need to prove the case $\gamma=2$ since the cases $\gamma=1$ and $0$ are bounded by it. We choose $f(x)=x^2e^{|x|}$, and the goal is to upper bound $\E f(Z)$ for $Z \sim \mathcal{P}_{\eta}$ by some polynomial rate of $1/\eta$. The dominated function is $g(x)=x^2e^{2|x|}$, and the set is chosen as 
$$\mathcal{A}_\eta=\{\tilde{x}_k:|k| < \left \lceil q\log(1/\eta)/\sqrt{\eta}\right \rceil\}.$$
It is clear that $g(\tilde{x}_k)\ge f(\tilde{x}_k)$ when $|k|\ge  \left\lceil q\log(1/\eta)/\sqrt{\eta}\right \rceil$ for small $\eta$. Once we show $$\E [g(Z_1) \mathbbm{1}_{ Z_1 \in \mathcal{A}_{\eta}^c}\mid Z_0=z] \leq g(z)-f(z)$$ 
for all $z \in \mathcal{A}_\eta^c$, we can use Lemma \ref{tweedie} to bound $\E f(Z)$.

\textbf{Step 3.}\label{Lstep3} We directly analyze $\E [g(Z_1) \mid Z_0=z]$ for $z \in \mathcal{A}_\eta^c$. Since the transition probability is known, we explicitly compute this conditional expectation and use Taylor expansion on $F$, the cumulative function, to upper bound $\E [g(Z_1) \mid Z_0=z]$ by $g(z)-f(z)$. Details of Steps 2 and 3 can be found in Section \ref{PLemma}.

\section{Discussion}
In this paper, we thoroughly studied the online quantile estimation and inference with the constant learning-rate SGD, which is a non-smooth and non-strongly-convex problem. Leveraging tools from Markov chain theory and the characteristic function, we showed that the unique stationary distribution of SGD iterations for the quantile loss is $\sqrt{\eta}-$asymptotically normal with minimal assumptions. It is one of the first CLT-type results for constant learning-rate stochastic approximation under the non-smooth setting. To achieve this goal, we established the convergence theorem of the periodic Markov chain induced by SGD for the quantile loss. We further investigated the tail probability and moments of the stationary distribution, which demonstrated some concentration properties of this countable-state Markov chain. For the practical concern, we proposed the inference procedure and applied the fully online kernel density estimation for implementation, offering computational efficiency in consistency with the spirit of SGD. Simulation across various scenarios justified the validity of our theoretical conclusions and exhibited ideal empirical performance of online inference. 

There are several directions and extensions for future research. First, the methodology in this paper is potentially generalizable to other non-smooth or non-strongly-convex settings, such as quantile regression, robust regression, and geometric median estimation. %It may be of interest to bridge the gap between the countable-state Markov chain in this paper and the uncountable-state cases. 
Moreover, the CLT in this paper does not have an explicit convergence rate. To remedy this limitation, we can consider deriving a Gaussian approximation result for quantile loss SGD, which can also enable practitioners to construct asymptotically pivotal statistics and confidence sets with non-asymptotic guarantees for powerful statistical inference.

\appendices
\section{Proof of Proposition \ref{markovchain}}\label{secA}
\begin{proof}
The Markov chain $\{\theta_{n}\}$ is irreducible as discussed before. It suffices to show the positive recurrence. We use Lemma \ref{foster} to prove it. Denote $\Theta$ as the state space. Without loss of generality, we assume $p<q/2$. The case $p>q/2$ can be proved by a similar argument. The case $p=q/2$ reduces to $\tau=1/2$, where we already argued that the Markov chain is positive recurrent and derived a closed-form stationary distribution.

Define $L(x)=| (x-x_0) q / \eta|+1$. Then $L(x_0+k\eta/q)=|k|+1$. Let $N_{1} = \min_{k} \{ F(x_0+k\eta/q) >2p/q  \}$. %Notice that $N_1\geq0$ since $F(x_0-\eta/q) \leq p/q $. 
When $L(\theta_{n-1}) > |N_{1}|+q+1$ and $\theta_{n-1}>\theta(p/q)$, we have
\begin{equation}\label{drift1}
    \mathbb{E}[L(\theta_n) - L(\theta_{n-1}) \mid \theta_{n-1}] < -(q-p)\frac{2p}{q}+\frac{p(q-2p)}{q}=-p.
\end{equation}
Similarly, we can choose $N_{2} = \max_{k} \{ F(x_0+k\eta/q) <p/(2q) \} \leq 0$. When $L(\theta_{n-1}) > |N_{2}|+q+1$ and $\theta_{n-1}<\theta(p/q)$, we have 
\begin{equation}\label{drift2}
\mathbb{E}[L(\theta_n) - L(\theta_{n-1}) \mid \theta_{n-1}] < (q-p)\frac{p}{2q}-\frac{p(2q-p)}{2q}=-\frac{p}{2}.
\end{equation}
So we can choose $N=\max\{|N_{1}|,|N_{2}|\}+q+1$ and $\epsilon=p/2$. Let $\mathcal{F}= \{ \theta \in \Theta  : L(\theta) \leq N \}$ which is finite for fixed $\eta$. We also have $L(Z)<\infty$ when $Z \in \mathcal{F}$. The last drift condition of Lemma \ref{foster} is verified by inequalities \eqref{drift1}-\eqref{drift2}. As a result, we have proved that the Markov chain is positive recurrent.
\end{proof}

\section{Proof of Lemma \ref{MGFbound}}\label{PLemma}

\begin{proof}
Define two auxiliary functions $f(x)=|x|^{\gamma}e^{|x|}$ and $g(x)=|x|^{\gamma}e^{2|x|}$. We first prove the case when $\gamma=2$. Define $k_{\eta}=\left \lceil q\log(1/\eta)/\sqrt{\eta}\right \rceil$ and let $k \ge k_{\eta}$. We consider the expectation of $g(Z_{i+1}) \mid Z_i=\tilde{x}_k$,
\begin{align*}
&\quad\,\,\E [g(Z_{i+1}) \mid Z_i=\tilde{x}_k]\\
&=F_kg(\tilde{x}_{k-q+p})+(1-F_k)g(\tilde{x}_{k+p}) \\
&=F_k \tilde{x}^2_{k-q+p}e^{2\tilde{x}_k}e^{-\tfrac{2\sqrt{\eta}(q-p)}{q}}+(1-F_k)\tilde{x}_{k+p}^2e^{2\tilde{x}_k}e^{\tfrac{2\sqrt{\eta}p}{q}} \\
&=F_k \tilde{x}^2_{k}\big[1-\frac{2(q-p)}{k}+\frac{(q-p)^2}{k^2}\big]e^{2\tilde{x}_k}e^{-\tfrac{2\sqrt{\eta}(q-p)}{q}}\\
&\qquad+(1-F_k)\tilde{x}_{k}^2\big(1+\frac{2p}{k}+\frac{p^2}{k^2}\big)e^{2\tilde{x}_k}e^{\tfrac{2\sqrt{\eta}p}{q}} .
\end{align*}
By Taylor expansion of $e^x$ around $0$, 
$$e^{\tfrac{2\sqrt{\eta}p}{q}}=1+\frac{2p}{q}\sqrt{\eta}+\frac{2p^2}{q^2}\eta+\mathcal{O}(\eta^{1.5}),$$
$$e^{-\tfrac{2\sqrt{\eta}(q-p)}{q}}=1-\frac{2(q-p)}{q}\sqrt{\eta}+\frac{2(q-p)^2}{q^2}\eta+\mathcal{O}(\eta^{1.5}).$$
We have the following bound,
\begin{align}
& \quad\,\,\frac{1}{g(\tilde{x}_k)}\E [g(Z_{i+1}) \mid Z_i=\tilde{x}_k]-1+\frac{f(\tilde{x}_k)}{g(\tilde{x}_k)}\nonumber\\
&=F_k\big[1-\frac{2(q-p)}{k}+\frac{(q-p)^2}{k^2}\big]e^{-\tfrac{2\sqrt{\eta}(q-p)}{q}}\\
&\qquad+(1-F_k)\big(1+\frac{2p}{k}+\frac{p^2}{k^2}\big)e^{\tfrac{2\sqrt{\eta}p}{q}}+e^{-\tilde{x}_k}\nonumber\\
&=-(F_k-\frac{p}{q})\frac{2q}{k}-2(F_k-\frac{p}{q})\sqrt{\eta}+\mathcal{O}(\eta)\nonumber\\
&\leq -2(F_k-\frac{p}{q})\sqrt{\eta}+\mathcal{O}(\eta).\label{mgf2}
\end{align}
Here and in the sequel, we use $d_0$ and $d_1$ to denote the probability density and its derivative at the true quantile, i.e., $d_0=f_X(\theta(\tau))$ and $d_1=f_X'(\theta(\tau))$. By Taylor expansion of $F(x)$ around $\theta(\tau)$ 
$$F_{k_{\eta}}=\frac{p}{q}+d_0(\frac{\eta  k_{\eta}}{q}+x_0-\theta(\tau))+\mathcal{O}(\eta ^2k_{\eta}^2).$$
Notice that $F_k$ is increasing in $k$, 
$$F_k -\frac{p}{q} \ge F_{k_{\eta}} -\frac{p}{q}=\frac{\eta d_0 k_{\eta}}{q}+\mathcal{O}(\eta ^2k_{\eta}^2).$$
Plug this into inequality \eqref{mgf2}, 
\begin{align*}
    &\quad\,\,\frac{1}{g(\tilde{x}_k)}\E [g(Z_{i+1}) \mid Z_i=\tilde{x}_k]-1+\frac{f(\tilde{x}_k)}{g(\tilde{x}_k)}\\
    &\leq -\frac{2d_0k_{\eta}\eta^{1.5}}{q}+\mathcal{O}(\eta). %\\
    %&\lesssim \eta \log(\eta). 
\end{align*}
Here $2d_0k_{\eta}\eta^{1.5}/q = \mathcal{O}(\eta\log(1/\eta))$. Hence for all $\eta$ sufficiently small and any $k \ge k_\eta$, the right hand side above is smaller than $0$, and we have $\E [g(Z_{i+1}) \mid Z_i=\tilde{x}_k] \leq g(\tilde{x}_k)-f(\tilde{x}_k)$. The same result can be identically proved for $k \leq -k_{\eta}$ since $f$ and $g$ are even functions. Moreover, notice that when $Z_i=\tilde{x}_k$ with $|k| \leq k_{\eta}$, we must have $|Z_{i+1}| \leq k_{\eta}\sqrt{\eta}/q+\sqrt{\eta}$ because the Markov chain can move by at most $\sqrt{\eta}$ at each step. Therefore, the following bound holds,
%it is clear that 
%$$\sup_{-k_{\eta} \leq k \leq k_{\eta}}\{g(\tilde{x}_k)-f(\tilde{x}_k)\}\leq g(\tilde{x}_{k_\eta})\lesssim \frac{\log(\eta)^2}{\eta^2}. $$
\begin{align*}
    \sup_{-k_{\eta} \leq k \leq k_{\eta}} \E [g(Z_{i+1}) \mid Z_i=\tilde{x}_k] &\leq g\Big(\frac{k_{\eta} \sqrt{\eta} }{q}+\sqrt{\eta}\Big) \\
    &\leq 2\Big( \log(1/\eta)+\sqrt{\eta}\Big)^2\frac{1}{\eta^2}\\ &\lesssim \frac{\log(1/\eta)^2}{\eta^2}.
\end{align*}
Let $\{Z \in \mathcal{A}_{\eta}\}$ denote the event that $Z= \tilde{x}_k$ for some $-k_{\eta} \leq k \leq k_{\eta}$. By Lemma \ref{tweedie}, we can bound the expectation of $f(Z)$ under the stationary distribution by
\begin{align*}
 \E f(Z) &= \E  f(Z)\mathbbm{1}_{ Z \in \mathcal{A}_{\eta}}+\E f(Z)\mathbbm{1}_{ Z \in \mathcal{A}_{\eta}^c} \\
 &\leq \sup_{ Z \in \mathcal{A}_{\eta}} f(Z)+ \sup_{z \in \mathcal{A}_{\eta}} \E[g(Z_1)\mathbbm{1}_{Z_1 \in \mathcal{A}_\eta^c}\mid Z_0=z].   
\end{align*}
It is also clear that  $\sup_{ Z \in \mathcal{A}_{\eta}} f(Z) \lesssim (\log\eta)^2/\eta^2$. So we can conclude that for any $\beta>3$, $\E_{Z \sim \mathcal{P}_{\eta}} f(Z) \leq \eta^{1-\beta}$ for all $\eta$ sufficiently small. In other words,
$$\sum_{k \in \mathbb{Z}} \eta^{\beta} \pi_k k^2e^{\frac{|k|\sqrt{\eta}}{q}} \leq q^2,$$
which completes the proof of the case when $\gamma=2$. The conclusion for $\gamma=1$ and $0$ follows immediately as they are bounded by the case $\gamma=2$.
\end{proof}

\section{Proof of Corollary \ref{tail}}

\begin{proof}
Since $\exp(N\sqrt{\eta}/q)\ge \exp(K_0\log(1/\eta))\ge \eta^{-K_0}$, we have
\begin{align*}
\sum_{|k| \ge N} \eta^{\beta-K_0} \pi_{\eta,k} |k|^{d} &\leq \sum_{|k| \ge N} \eta^{\beta} \pi_{\eta,k} |k|^{d}e^{\frac{N\sqrt{\eta}}{q}} \\
&\leq \sum_{k \in \mathbb{Z}} \eta^{\beta} \pi_{\eta,k} |k|^{d}e^{\frac{|k|\sqrt{\eta}}{q}} \\
&\leq q^2.
\end{align*}
\end{proof}

\section{Proof of Corollary \ref{Conineq}}

\begin{proof}
Let $Z$ follow the stationary distribution of $\{ \theta_n(\tau) \}$. By definition, we have $|x_0 - \theta(\tau)| \leq \eta/q$. Then by Lemma \ref{MGFbound} and Markov inequality, for any $\epsilon >\eta/q$,
\begin{align*}
    &\quad\,\,\P (|Z-\theta(\tau)|  \ge \epsilon) \\
    &\leq \P \Big(|Z-x_0|  \ge \epsilon - \frac{\eta}{q}\Big) \\
    & = \P \Big(\frac{|Z-x_0|}{\sqrt{\eta}}  \ge \frac{\epsilon}{\sqrt{\eta}} -\frac{\sqrt{\eta}}{q}\Big)\\
    &= \P \Big(  \frac{(Z-x_0)^2}{\eta} \exp (\frac{|Z-x_0|}{\sqrt{\eta}}) \\
    &\qquad\qquad\qquad \ge  (\frac{\epsilon}{\sqrt{\eta}} -\frac{\sqrt{\eta}}{q})^2 \exp( \frac{\epsilon}{\sqrt{\eta}} -\frac{\sqrt{\eta}}{q} ) \Big) \\
    & \leq \frac{\E \{ \frac{(Z-x_0)^2}{\eta} \exp (\frac{|Z-x_0|}{\sqrt{\eta}}) \}}{ (\frac{\epsilon}{\sqrt{\eta}} -\frac{\sqrt{\eta}}{q})^2 \exp( \frac{\epsilon}{\sqrt{\eta}} -\frac{\sqrt{\eta}}{q} )} \\
    & \leq \eta^{1-\beta} (\frac{\epsilon}{\sqrt{\eta}} -\frac{\sqrt{\eta}}{q})^{-2}\exp( -\frac{\epsilon}{\sqrt{\eta}} +\frac{\sqrt{\eta}}{q}) \\
    &=\frac{ \eta^{2-\beta}} {\epsilon^2}\exp\Big( -\frac{\epsilon}{\sqrt{\eta}} +\frac{\sqrt{\eta}}{q}\Big) \frac{q^2}{q^2+\eta^2/\epsilon^2-2q \eta/\epsilon}.
\end{align*}
\end{proof}

\section{Proof of Theorem \ref{charac}}\label{secproofmain}
\begin{proof}
We first claim the following moment bound of the standardized stationary measure: for all $\eta$ sufficiently small, we have
\begin{equation}\label{moments}
    \E|Z| \leq K_1 \log(\frac{1}{\eta}), \quad  \E Z^2\leq K_2 (\log \eta)^2,
\end{equation}
	where $Z \sim \mathcal{P}_{\eta}$ and $K_1,K_2$ are some universal constants.

To prove the claim, let $\beta$ be the same as in Corollary \ref{tail}, $K_0=\beta+1$, and $N' = \lceil qK_0 \log (1/\eta)/ \sqrt{\eta} \rceil$. Then $\E (|Z|\mathbbm{1}_{|Z|<N'\sqrt{\eta}/q} ) \leq N'\sqrt{\eta}/q\lesssim\log(1/\eta).$ By Corollary \ref{tail}, $\E ( |Z|\mathbbm{1}_{|Z|\ge N'\sqrt{\eta}/q} )\lesssim \log(1/\eta)$ also holds. So the conclusion is proved. The same argument can be used to prove the second moment part.

Let $N = \left \lceil 5q \log (1/\eta)/ \sqrt{\eta}\right \rceil$. For any $t_0>0$, we investigate the relationship between $\phi_{\eta}(t)$ and its derivative on the interval $[0,t_0]$. We first require the learning rate $\eta \leq t_0^{-7}$. Let $Z$ denote the random variable following the standardized stationary distribution $\mathcal{P}_{\eta}$. We consider the characteristic function of $\mathcal{P}_{\eta}$, and plug in the equation for the stationary measure \eqref{station}:
\begin{align*}
	\phi_{\eta}(t)&=\E e^{itZ} \\
	&=\sum_{k=-\infty}^{\infty} \pi_k e^{\tfrac{itk\sqrt{\eta}}{q}} \\
	&=\sum_{k=-\infty}^{\infty} (\pi_{k+q-p}F_{k+q-p}+ \pi_{k-p}(1-F_{k-p}))e^{\tfrac{itk\sqrt{\eta}}{q}}\\
	&=\sum_{k=-N}^{N} (\pi_{k+q-p}F_{k+q-p}+ \pi_{k-p}(1-F_{k-p}))e^{\tfrac{itk\sqrt{\eta}}{q}} \\
    &\qquad\qquad\qquad+\mathcal{O}(\eta^2),
\end{align*}
where the last step is from $\sum_{|k|>N} (\pi_{k+q-p}F_{k+q-p}+ \pi_{k-p}(1-F_{k-p}))e^{\tfrac{itk\sqrt{\eta}}{q}}=\mathcal{O}(\eta^2)$ due to Corollary \ref{tail}. Here we have a truncated version of the characteristic function. Recall that $d_0$ and $d_1$ are the probability density and its derivative at the true quantile. By Taylor expansion of $F$ around the true quantile $\theta(\tau)$,
\begin{align*}&F_{k}=\frac{p}{q} +\frac{k\eta d_0}{q}+(x_0-\theta(\tau))d_0\\
    &\qquad\qquad\qquad+(\frac{k\eta}{q}+x_0-\theta(\tau))^2\frac{d_1}{2}+\mathcal{O}(k^3\eta^3),
\end{align*}
where $|x_0-\theta(\tau)|=\mathcal{O}(\eta)$ is fixed. We plug this into the two terms of the truncated formula of the characteristic function and get
$$\sum_{k=-N}^{N} \pi_{k+q-p}F_{k+q-p}e^{\tfrac{itk\sqrt{\eta}}{q}}=\text{I}_0+\text{I}_1+\text{I}_2+\mathcal{O}(k^3\eta^3),$$
where
\begin{align*}
\text{I}_0=\sum_{k=-N}^{N} \pi_{k+q-p}e^{\tfrac{itk\sqrt{\eta}}{q}} \Big[ &\frac{p}{q}+(x_0-\theta(\tau))d_0 \\
&+\frac{d_1(x_0-\theta(\tau))^2}{2}\Big],    
\end{align*}
$$\text{I}_1=\sum_{k=-N}^{N} \pi_{k+q-p}e^{\tfrac{itk\sqrt{\eta}}{q}}\frac{(k+q-p)\eta [d_0+(x_0-\theta(\tau))d_1]}{q} ,$$
$$\text{I}_2=\sum_{k=-N}^{N} \pi_{k+q-p}e^{\tfrac{itk\sqrt{\eta}}{q}}\frac{(k+q-p)^2\eta^2d_1}{2q^2},$$
and the remainder term is from Taylor expansion.
Similarly for the other term:
$$\sum_{k=-N}^{N} \pi_{k-p}(1-F_{k-p})e^{\tfrac{itk\sqrt{\eta}}{q}}=\II_0+\II_1+\II_2+\mathcal{O}(k^3\eta^3),$$
where
\begin{align*}\II_0=\sum_{k=-N}^{N} \pi_{k-p}e^{\tfrac{itk\sqrt{\eta}}{q}}\Big[ \frac{q-p}{q}&-(x_0-\theta(\tau))d_0\\
&-\frac{d_1(x_0-\theta(\tau))^2}{2}\Big],
\end{align*}
$$\II_1=-\sum_{k=-N}^{N} \pi_{k-p}e^{\tfrac{itk\sqrt{\eta}}{q}}\frac{(k-p)\eta [d_0+(x_0-\theta(\tau))d_1]}{q},$$
$$\II_2=-\sum_{k=-N}^{N} \pi_{k-p}e^{\tfrac{itk\sqrt{\eta}}{q}}\frac{(k-p)^2\eta^2d_1}{2q^2}.$$
Now we apply variable shift to get the following relationship,
\begin{align}\label{shift1}
	&\quad\,\,\sum_{k=-N}^{N} \pi_{k+q-p}e^{\tfrac{itk\sqrt{\eta}}{q}}\\
    &=e^{\tfrac{it(p-q)\sqrt{\eta}}{q}}\sum_{k=-N}^{N}  \pi_{k+q-p}e^{\tfrac{it(k+q-p)\sqrt{\eta}}{q}} \nonumber \\
		&=\phi_{\eta}(t)e^{\tfrac{it(p-q)\sqrt{\eta}}{q}}+\mathcal{O}(\eta^2),
\end{align}
where the order of the remainder $\mathcal{O}(\eta^2)$ is from Corollary \ref{tail}. We also have 
\begin{align*} 
	&\quad\,\, \phi'_{\eta} (t)\\
    &=\sum_{k=-\infty}^{\infty} \frac{ik\sqrt{\eta}}{q} \pi_k e^{\tfrac{itk\sqrt{\eta}}{q}} \\
	&= \sum_{k=-\infty}^{\infty} \frac{i(k+q-p)\sqrt{\eta}}{q} \pi_{k+q-p} e^{\tfrac{it(k+q-p)\sqrt{\eta}}{q}}\\
	&= \sum_{k=-N}^{N} \frac{i(k+q-p)\sqrt{\eta}}{q} \pi_{k+q-p} e^{\tfrac{it(k+q-p)\sqrt{\eta}}{q}}+\mathcal{O}(\eta^{2}),
\end{align*}
and as a result,
\begin{align}\label{shift2}
    &\quad\,\, \sum_{k=-N}^{N} \frac{i\sqrt{\eta}}{q} \pi_{k+q-p}(k+q-p)e^{\tfrac{itk\sqrt{\eta}}{q}}\\
    &=\phi'_{\eta}(t)e^{\tfrac{it(p-q)\sqrt{\eta}}{q}}+\mathcal{O}(\eta^{2}).
\end{align}
Similarly,
\begin{align}\label{shift3}
    &\quad\,\, -\sum_{k=-N}^{N} \frac{\eta}{q^2} \pi_{k+q-p}(k+q-p)^2e^{\tfrac{itk\sqrt{\eta}}{q}}\\
    &=\phi''_{\eta}(t)e^{\tfrac{it(p-q)\sqrt{\eta}}{q}}+\mathcal{O}(\eta^{2}).
\end{align}
Now we can plug equations \eqref{shift1}-\eqref{shift3} into the formula of $\text{I}_0$, $\text{I}_1$ and $\text{I}_2$:
\begin{align*}
\text{I}_0=\phi_{\eta}(t)e^{\tfrac{it(p-q)\sqrt{\eta}}{q}}\big[ \frac{p}{q}&+(x_0-\theta(\tau))d_0\\
&+\frac{d_1(x_0-\theta(\tau))^2}{2}\big]+\mathcal{O}(\eta^2),    
\end{align*}
$$\text{I}_1=-i\phi'_{\eta}(t)e^{\tfrac{it(p-q)\sqrt{\eta}}{q}}\sqrt{\eta} [d_0+(x_0-\theta(\tau))d_1]+\mathcal{O}(\eta^{2}),$$
$$\text{I}_2=-\phi''_{\eta}(t)e^{\tfrac{it(p-q)\sqrt{\eta}}{q}}\frac{\eta d_1}{2}+\mathcal{O}(\eta^{2}).$$
The same argument works for the second part,
$$\sum_{k=-N}^{N} \pi_{k-p}e^{\tfrac{itk\sqrt{\eta}}{q}}=\phi_{\eta}(t)e^{\tfrac{itp\sqrt{\eta}}{q}}+\mathcal{O}(\eta^2),$$
$$\sum_{k=-N}^{N} \frac{i\sqrt{\eta}}{q} \pi_{k-p}(k-p)e^{\tfrac{itk\sqrt{\eta}}{q}}=\phi'_{\eta}(t)e^{\tfrac{itp\sqrt{\eta}}{q}}+\mathcal{O}(\eta^{2}),$$
$$-\sum_{k=-N}^{N} \frac{\eta}{q^2} \pi_{k-p}(k-p)^2e^{\tfrac{itk\sqrt{\eta}}{q}}=\phi''_{\eta}(t)e^{\tfrac{itp\sqrt{\eta}}{q}}+\mathcal{O}(\eta^{2}),$$
and hence
\begin{align*}
\II_0=\phi_{\eta}(t)e^{\tfrac{itp\sqrt{\eta}}{q}}\big[ \frac{q-p}{q}&-(x_0-\theta(\tau))d_0
\\&-\frac{d_1(x_0-\theta(\tau))^2}{2}\big]+\mathcal{O}(\eta^2),    
\end{align*}
$$\II_1=i\phi'_{\eta}(t)e^{\tfrac{itp\sqrt{\eta}}{q}}\sqrt{\eta}[d_0+(x_0-\theta(\tau))d_1]+\mathcal{O}(\eta^{2}),$$
$$\II_2=\phi''_{\eta}(t)e^{\tfrac{itp\sqrt{\eta}}{q}}\frac{\eta d_1}{2}+\mathcal{O}(\eta^{2}).$$
Thereby
$$\phi_{\eta}(t) = \text I_0+\II_0+\text{I}_1+\II_1+\text{I}_2+\II_2+\mathcal{O}\big((-\log\eta)^3\eta^{1.5}\big).$$
The order of the remainder is due to $k^3\eta^3\leq N^3\eta^3\asymp(-\log\eta)^3\eta^{1.5}.$
We now sum them up correspondingly, using the following Taylor expansion:
$$e^{\tfrac{p\sqrt{\eta}it}{q}}=1+\frac{p\sqrt{\eta}it}{q}-\frac{\eta p^2 t^2}{2q^2}+\mathcal{O}(\eta^{1.5}),$$
$$e^{-(q-p)\tfrac{\sqrt{\eta}it}{q}}=1-\frac{(q-p)\sqrt{\eta}it}{q}-\frac{(q-p)^2\eta t^2}{2q^2}+\mathcal{O}(\eta^{1.5}).$$
Recall that $t<t_0$ is bounded, so the remainder term does not include $t$. We first deal with $\text{I}_0+\II_0-\phi_{\eta}(t)$, all the terms related with $\phi_\eta(t)$. After Taylor expansion on the exponential term, the coefficient on $\phi_\eta(t)$ becomes
\begin{align*} &\quad\,\,\Big(\frac{p}{q}+(x_0-\theta(\tau))d_0+\frac{d_1(x_0-\theta(\tau))^2}{2}\Big)\\
&\qquad\qquad\cdot\Big(1-\frac{(q-p)\sqrt{\eta}it}{q}-\frac{(q-p)^2\eta t^2}{2q^2}\Big)\\
&\qquad+ \Big(\frac{q-p}{q}-(x_0-\theta(\tau))d_0-\frac{d_1(x_0-\theta(\tau))^2}{2}\Big)\\
&\qquad\qquad\cdot\Big(1+\frac{p\sqrt{\eta}it}{q}-\frac{\eta p^2 t^2}{2q^2}\Big)-1+\mathcal{O}(\eta^{1.5})\\
& = \Big(-\frac{p}{q}\frac{(q-p)^2\eta t^2}{2q^2}-\frac{(q-p)}{q}\frac{\eta p^2 t^2}{2q^2}\Big)+\mathcal{O}(\eta^{1.5})\\
&= -\frac{p(q-p)\eta t^2}{2q^2} + \mathcal{O}(\eta^{1.5}).
\end{align*}
So we have 
$$\text{I}_0+\II_0-\phi_{\eta}(t)=-\frac{p(q-p)\eta t^2}{2q^2}\phi_{\eta}(t)+\mathcal{O}(\eta^{1.5}).$$
Similarly for $\text{I}_1+\II_1$, notice that \eqref{moments} implies $|\phi_{\eta}'(t)|\lesssim  \log(\eta^{-1})$, the coefficient on $\phi_{\eta}'(t)$ becomes
\begin{align*} &\quad\,\,\sqrt{\eta}i[d_0+(x_0-\theta(\tau))d_1]\\
&\qquad\qquad\cdot\Big(1+\frac{p\sqrt{\eta}it}{q}-1+\frac{(q-p)\sqrt{\eta}it}{q}\Big)+\mathcal{O}(\eta^{1.5})\\
&=-\eta d_0t+\mathcal{O}(\eta^{1.5}),
\end{align*}
which leads to 
$$\text{I}_1+\II_1=-\eta d_0t\phi_{\eta}'(t)+\mathcal{O}(\log(\eta^{-1})\eta^{1.5}).$$
By \eqref{moments},  $|\phi_{\eta}''(t)|\lesssim  (\log \eta)^2$. So $\text{I}_2+\II_2 \lesssim \mathcal{O}((\log\eta)^2\eta^{1.5}) \lesssim \mathcal{O}\big((-\log\eta)^3\eta^{1.5}\big)$. We finally get the following result,
\begin{align*}
R_{\eta}(t) &:=\frac{p(q-p)\eta t^2}{2q^2}\phi_{\eta}(t)+\eta d_0t\phi_{\eta}'(t)\\
&=\mathcal{O}\big((-\log\eta)^3\eta^{1.5}\big).    
\end{align*}
Define $$D_{\eta}(t)=\exp\big(\frac{p(q-p) t^2}{4q^2d_0}\big)\phi_{\eta}(t),$$
with the derivative
\begin{align}\label{deri}
    D'_{\eta}(t)&=\exp\big(\frac{p(q-p) t^2}{4q^2d_0}\big)\phi'_{\eta}(t)\nonumber\\
    &\qquad\qquad+\frac{p(q-p) t}{2q^2d_0}\exp\big(\frac{p(q-p) t^2}{4q^2d_0}\big)\phi_{\eta}(t)\nonumber\\
    &=\exp\big(\frac{p(q-p) t^2}{4q^2d_0}\big)\frac{R_{\eta}(t)}{\eta d_0 t}.
\end{align}
For any $t_0>0$, the previous argument showed that there exists a universal constant $C$ such that
$$ |t D'_{\eta}(t)|\leq C \exp\big(\frac{p(q-p) t_0^2}{4q^2d_0}\big)(-\log\eta)^3\sqrt{\eta}$$
for all $t \in [0,t_0]$. Choose $\delta_{\eta}=-\eta^{\tfrac{1}{4}}\log \eta$, the following bound holds,
$$ \sup_{t \in [\delta_{\eta}, t_0]} |D'_{\eta}(t)|\leq C_{t_0}  (\log\eta)^2 \eta^{\tfrac{1}{4}},$$
where $C_{t_0}=\exp\big({p(q-p) t_0^2}/(4q^2d_0)\big)$.
Moreover we can bound the derivative of $D_{\eta}$ on $[0,\delta_{\eta}]$ by \eqref{deri} as 
\begin{align*}
\sup_{t \in [0,\delta_{\eta}]}|D'_{\eta}(t)| 
&\leq C_{t_0} \sup_{t \in [0,\delta_{\eta}]}|\phi_{\eta}'(t)| 
+ \frac{p(q-p)t_0}{2q^2d_0} C_{t_0} \\
&\leq 2 C_{t_0} \E_{Z \sim \mathcal{P}_{\eta}}|Z|
+ \frac{p(q-p)t_0}{2q^2d_0}  C_{t_0} \\
&\leq K_{t_0} \log\Big(\frac{1}{\eta}\Big).
\end{align*}
where $K_{t_0}$ is another constant only depended on $t_0$.

Finally, by the fundamental theorem of calculus, we have 
\begin{align*}
&\quad\,\, |D_{\eta}(t_0)-D_{\eta}(0)| \\
&\leq \delta_{\eta}\sup_{t \in [0,\delta_{\eta}]}|D'_{\eta}(t)|
+(t_0-\delta_{\eta})\sup_{t \in [\delta_{\eta}, t_0]}|D'_{\eta}(t)| \\
&\lesssim (\log\eta)^2 \eta^{\tfrac{1}{4}} \rightarrow 0
\end{align*}
as $\eta \rightarrow 0$. The identical argument can be used to prove the case when $t_0 <0$. Since $D_{\eta}(0)=1$, we have proved that the pointwise convergence $D_{\eta}(t_0)\rightarrow 1$ holds for any $t_0 \in \mathbb{R}$. Equivalently, for any $t \in \mathbb{R}$,
$$ \lim_{\eta \rightarrow 0} \phi_{\eta}(t)=e^{-\tfrac{p(q-p)t^2}{4q^2d_0}}.$$
\end{proof}

\section{Proof of Theorem \ref{OKDE}}

\begin{proof}
Consider the online kernel density estimator
\begin{equation}
    \hat{f}_n(\tau) = \frac{1}{n} \sum_{k=1}^{n} \frac{K_{b_k}(\theta_{k-1}(\tau), X_k)}{b_k},
\end{equation}
where the bandwidth $b_k \asymp k^{-\alpha}$ for some $0 <\alpha <1$. We first show that with probability $1$, $\hat{f}_n(\tau) \rightarrow \E f_X(\theta_{\infty}(\tau))$. To this end, define $V_n = \frac{1}{n}\sum_{i=1}^n f_X(\theta_{i-1}(\tau))$ and
\begin{align*}
W_n &= \frac{1}{n} \sum_{i=1}^n \E \Big\{ \frac{K_{b_i}(\theta_{i-1}(\tau), X_i)}{b_i} \Big|  \mathcal{F}_{i-1}\Big\} \\
&= \frac{1}{n} \sum_{i=1}^n \int_{-M}^{M}  K (u) f_X(\theta_{i-1}(\tau)-b_i u)du  .    
\end{align*}
Since $n \hat{f}_n(\tau)-nW_n$ is a sum of martingale differences, by Burkholder's inequality,
$$\E \max_{k \leq n} (k \hat{f}_k(\tau)-kW_k)^2 \lesssim \sum_{i=1}^n \frac{1}{b_i} = n^{1+\alpha}.$$
By Taylor expansion,
\begin{align*}
&\quad\,\,f_X(\theta_{i-1}(\tau)-b_i u)-f_X(\theta_{i-1}(\tau))+b_i u f_X'(\theta_{i-1}(\tau)) \\
&= \mathcal{O}(b^2_i u^2),    
\end{align*}
thereby 
$$|W_n-V_n|\lesssim \frac{1}{n} \sum_{i=1}^nb_i^2 \rightarrow 0.$$
Since we have shown that $\{\theta_n(\tau)\}$ is positive recurrent, by the ergodic theorem, $V_n \rightarrow \E f_X(\theta_{\infty}(\tau))$ almost surely. Now we have proved that the estimator is consistent for $\E f_X(\theta_{\infty}(\tau))$, and it suffices to bound the difference between $\E f_X(\theta_{\infty}(\tau))$ and $f_X(\theta(\tau))$. Since $\E | \theta_{\infty}(\tau)-\theta(\tau)| \lesssim \sqrt{\eta}\log(1/\eta)$ and $|f'_X|$ is bounded, we have
\begin{align*} 
|\E f_X(\theta_{\infty}(\tau)) - f_X(\theta(\tau))| &\leq \E | f_X(\theta_{\infty}(\tau)) - f_X(\theta(\tau))| \\
&\lesssim \E |\theta_{\infty}(\tau)-\theta(\tau)| \\
&\lesssim \sqrt{\eta}\log(1/\eta).
\end{align*}
So $\E f_X(\theta_{\infty}(\tau)) \rightarrow f_X(\theta(\tau)) $ as $\eta \rightarrow 0$.

\end{proof}

\section{Discussion on the Joint CLT}\label{sec_joint_CLT}

While the previous sections focused on univariate SGD quantile estimates for clarity, our methodology extends to the multivariate case through a more sophisticated treatment. In particular, we can derive the joint CLT for the multivariate quantile estimation which includes interaction between coordinates. Let $\db{X}_j=(X_{j ,1}, \dots, X_{j,d})^\top$ be i.i.d. $\R^d$-valued random vectors with a joint distribution function $F_{\db{X}}$ and density $f_{\db{X}}$. With a slight abuse of notation we write $\db{X}_1=(X_{ 1}, \dots, X_{ d})^\top$. Let $\theta_i(\tau_i)$ be the $\tau_i$th quantile of $X_i$, $0 < \tau_i < 1$. Our goal is to estimate the quantiles vector
\[
\boldsymbol{\theta}^{*}=(\theta_1^{*},\ldots,\theta_d^{*})^\top\in\R^d, 
\]
with quantile levels $\tau_i=p_i/q_i\in(0,1)$ where $p_i$ and $q_i$ are coprime integers. Here we use $\theta_i^*$ to denote $\theta_i(\tau_i)$ for notational simplicity. The objective function for this problem is
\begin{align}
\boldsymbol{\theta}^*=\mathop{\mathrm{arg\,min}}_{\boldsymbol{\theta} = (\theta_1,\ldots,\theta_d)^\top \in \mathbb{R}^d} \sum_{i=1}^d \E\{(X_i-\theta_i)(\tau_i - \mathbbm{1}_{\theta_i\ge X_i})\},  
\end{align}
and the SGD recursion can be written as the coordinate-wise update for the quantile estimation SGD (with the same learning rate $\eta$). Let $\db{\theta}_n=(\theta_{n,1},\ldots, \theta_{n,d} )^{\top}$, the coordinate-wise SGD iterates are
\begin{align}
	\theta_{n+1,j}=\theta_{n,j}+\eta [ \tau_j \mathds{1}_{X_{n+1,j}>\theta_{n,j}} -(1-\tau_j) \mathds{1}_{X_{n+1,j}\leq \theta_{n,j}} ]
\end{align}
for $1\leq j \leq d$. We have already established the limiting law for the marginal distributions. To study the joint distribution, it suffices to leverage the same method in our paper, and the joint CLT with closed-form limiting covariance can be obtained.

To this end, consider this $d$-dimensional SGD as a multivariate Markov chain. The centered (at $\db{\theta}^*$) and standardized (by $\sqrt{\eta}$) multivariate state space is
$$ \boldsymbol{\theta} = (\theta_1,\dots,\theta_d)^\top \in  \db{\mathcal{M}} =  \mathcal{M}_1(\tau_1)\times \cdots \times \mathcal{M}_d(\tau_d),$$
where
$$ \mathcal{M}_i(\tau_i)=\Big\{ \theta_i = x_i+ \frac{k\sqrt{\eta}}{q_i}, \ \ k \in \mathbb{Z}\Big\}.$$
Here $\{x_i\}_{i=1}^d$ are some nuisance terms due to initialization. Given any $\boldsymbol{\theta} \in \R^d$ in the state space, there are $2^d$ possible previous states (or future states) since each coordinate can either move forward or backward. The following theorem generalizes the main result to the multivariate case with the explicit limiting covariance matrix.

\begin{theorem}
Suppose that the random vector $\db{X}$ has a bounded density function $f_{\db{X}}$ being $C^2$ smooth in an $r$-neighborhood of $\db{\theta}^*$: $\db{\mathcal{B}}_r(\db{\theta}^*)= \{\db{\theta} \in \R^d: \| \db{\theta} -\db{\theta}^*\| \leq r\}$, for some $r>0$, and $f_{\db{X}}(\db{\theta}^*)>0$. For $\eta$ sufficiently small, the Markov chain of the multivariate quantile SGD has a unique stationary distribution $\db{m}_\eta(\cdot)$. Let $\db{\mathcal{P}}_{\eta} (\cdot) = \db{m}_\eta(\db{\theta}^* + \sqrt \eta\  \cdot)$ be the centered and standardized measure. We have
    $$\db{\mathcal{P}}_{\eta} \stackrel{\mathcal{D}}{\to} \mathcal{N}(0, V),\quad \text{as }\eta\rightarrow0,$$
where the closed-form covariance matrix $V$ is 
$$ V_{ij} = \frac{\mathbb{P}(X_i \leq \theta^*_i, X_j \leq \theta^*_j) -\tau_{i}\tau_{j}}{f_{X_i}(\theta^*_i)+f_{X_j}(\theta^*_j)} .$$
\end{theorem}

\begin{proof} 
For the existence and uniqueness of a stationary distribution for the multivariate case, we shall use the Lyapunov function in Proposition 1 in \cite{zhang_piecewise_2025}. Specifically, according to the latter result, there exists positive constants $\Delta$, $\alpha_0$, $\mu_0$ and $c_0$, such that the Lyapunov function $L(\db{\theta}) = \max( e^{-1} 2 \exp(R(\db{\theta})) - 1,  R(\db{\theta})^2)$, where $R(\db{\theta}) = \|\db{\theta} - \db{\theta}^*\| / \Delta$, satisfies $\mathbb{E}\!\left[ L\!\bigl(\db\theta _{n+1}\bigr) | \db\theta _{n} \right] \le (1 - \eta \mu_{0}) L(\db\theta_n) + \eta^2 c_{0}$ for all $0 < \eta\leq \alpha_0$. This Lyapunov function is defined on the original SGD sequence. Thus, by Lemma \ref{foster}, for such $\eta$ the process is positive recurrent and thus has a stationary solution.

We then define $\boldsymbol{l}=(l_1,\dots,l_d)\in\{0,1\}^d$ to indicate the movement for each coordinate. For some current value of the SGD iterate $\db{u}=(u_1,\dots,u_d)^\top$ and the data $\boldsymbol{X} = (X_1,\dots,X_d)^\top$, let $l_i = \ind_{X_i \leq u_i}$. So $l_i=1$ means the $i$-th coordinate of the Markov chain moves backward and vice versa. Further define the orthant event
\begin{equation}
E_{\boldsymbol{l}}(\boldsymbol{u})
:=\Big(\bigcap_{i: \,l_i=1}\{X_i\le u_i\}\Big) \cap\
  \Big(\bigcap_{i: \,l_i=0}\{X_i> u_i\}\Big),
\end{equation}
and its probability
\begin{equation}
P_{\boldsymbol{l}}(\boldsymbol{u}):=\Pbb\big(E_{\boldsymbol{l}}(\boldsymbol{u})\big).
\label{eq:Ps}
\end{equation}
Define the drift vector $\db{v} (\db{l}) =  (v_1(\db{l}),\dots,v_d(\db{l}))^\top\in\R^d$ by
\begin{equation}
v_i(\db{l})=
\begin{cases}
\tau_i, & l_i=0,\\
\tau_i-1, & l_i=1,
\end{cases}
\qquad\Longleftrightarrow\qquad \db{v} (\db{l})=\db{\tau}-\db{l},
\end{equation}
where $\db{\tau}=(\tau_1,\dots,\tau_d)^\top$. 
Let $\pi_{{\db{\theta}}}$ denote the probability mass function of $\db{\theta}$ corresponding to $\db{\mathcal{P}}_{\eta}$ over the discrete state space $\db{\mathcal{M}}$. By the definition of the stationary measure, for any $\db{\theta} \in \db{\mathcal{M}}$ , we have
$$\pi_{\db{\theta}} = \sum_{\db{l} \in \{0,1\}^d} \pi_{\db{\theta}-\sqrt{\eta} \db{v} (\db{l})} P_{\db{l}}(\sqrt{\eta}\db{\theta}+\db{\theta}^*-\eta\db{v} (\db{l})).$$
Notice that the transition kernel is $P_{\db{l}}(\sqrt{\eta}\db{\theta}+\db{\theta}^*-\eta\db{v} (\db{l}))$ because we need to transform the previous states back to the original values of SGD iterates. For the characteristic function of the standardized Markov chain, we have
\begin{align}\label{eqch}
 &\quad \, \,\phi_{\eta}(\db{t})\nonumber\\
 &=\E_{\db{Z} \sim \pi} e^{i\db{t}^{\top}\db{Z}}\nonumber\\
 &=\sum_{\db{\theta} \in\db{\mathcal{M}} } \pi_{\db{\theta}} e^{i  \db{t}^{\top} \db{\theta}}\nonumber\\
 &=  \sum_{\db{\theta} \in\db{\mathcal{M}} } \Big[\sum_{\db{l} \in \{0,1\}^d} \pi_{\db{\theta}-\sqrt{\eta} \db{v} (\db{l})} P_{\db{l}}(\sqrt{\eta}\db{\theta}+\db{\theta}^*-\eta\db{v} (\db{l}))\Big] e^{i  \db{t}^{\top} \db{\theta}}.   
\end{align} 
Now the high-level idea is identical to the proof in Section \ref{secproofmain}. We interchange the summation $\sum_{\db{\theta}}$ and $\sum_{\db{l}}$, determine a cut-off in the sum $\sum_{\db{\theta} \in\db{\mathcal{M}} }$, and discard the tail. In particular, let $N = \lceil C \log(1/\eta)/\sqrt{\eta} \rceil$ for a sufficiently large constant $C$. We truncate the state space to a hypercube $\db{\mathcal{M}}_N = \db{\mathcal{M}} \cap [-N\sqrt{\eta}, N\sqrt{\eta}]^d$. The error introduced by this truncation is $\mathcal{O}(\eta^2)$ due to the exponential tail decay.

Then we leverage Taylor expansion of transition probability $P_{\db{l}}(\sqrt{\eta}\db{\theta}+\db{\theta}^*-\eta\db{v} (\db{l})) $ at $\db{\theta} = \db{\theta}^*$, 
\begin{align}\label{taylortran}
&\quad\,\, P_{\db{l}}(\sqrt{\eta}\db{\theta}+\db{\theta}^*-\eta\db{v} (\db{l}))\nonumber \\
&= P_{\db{l}}(\db{\theta}^*)
+ \sum_{k=1}^d \varphi_k(\db{l}) \,
\Big( \sqrt{\eta}\theta_k  - \eta\, v_k(\db{l}) \Big)
+\mathcal{R} \nonumber \\
& = P_{\db{l}}(\db{\theta}^*)
+ \db{\varphi}(\db{l})^\top \Big( \sqrt{\eta}\db{\theta}  - \eta\, \db{v}(\db{l}) \Big) +\mathcal{R},
\end{align}
where 
\begin{equation}
\varphi_k(\db{l}):=\left.\frac{\partial}{\partial \theta_k}P_{\db{l}}(\db{\theta})\right|_{\db{\theta}=\db{\theta}^{*}},
\qquad
\db{\varphi}(\db{l}):=(\varphi_1(\db{l}),\dots,\varphi_d(\db{l}))^\top,
\label{eq:phi_def}
\end{equation}
and $\mathcal{R}$ is the remainder. Now we have 
\begin{align}\label{updatedse}
 & \quad\,\, \phi_{\eta}(\db{t}) \nonumber\\
 &=  \sum_{\db{l}\in \{0,1\}^d} \Big\{ \sum_{\db{\theta} \in\db{\mathcal{M}}_N } \pi_{\db{\theta}-\sqrt{\eta} \db{v} (\db{l})} \nonumber\\
 &\qquad\qquad \cdot \Big[ P_{\db{l}}(\db{\theta}^*)
+ \sqrt{\eta} \db{\varphi}(\db{l})^\top \Big( \db{\theta}  - \sqrt{\eta}\, \db{v}(\db{l}) \Big) \Big] \Big\} e^{i  \db{t}^{\top} \db{\theta}} \nonumber\\
&\qquad+\mathcal{R} + \mathcal{O}(\eta^2). 
\end{align} 
Then we apply the variable shift (same as, e.g., \eqref{shift1}) to rewrite the right-hand side as a linear combination of $\phi_{\eta}(\db{t})$ and its derivative,
\begin{align} \label{vs1}
&\quad\,\, \sum_{\db{\theta} \in\db{\mathcal{M}}_N } \pi_{\db{\theta}-\sqrt{\eta} \db{v} (\db{l})}P_{\db{l}}(\db{\theta}^*) e^{i  \db{t}^{\top} \db{\theta}} \nonumber\\
&= P_{\db{l}}(\db{\theta}^*) \phi_{\eta}(\db{t}) e^{i\sqrt{\eta}\db{t}^{\top}\db{v} (\db{l}) } + \mathcal{O}(\eta^2),
\end{align} 
\begin{align}  \label{vs2}
&\quad\,\,\sum_{\db{\theta} \in\db{\mathcal{M}}_N } \pi_{\db{\theta}-\sqrt{\eta} \db{v} (\db{l})} \sqrt{\eta} \db{\varphi}(\db{l})^\top \Big( \db{\theta}  - \sqrt{\eta}\, \db{v}(\db{l}) \Big)e^{i  \db{t}^{\top} \db{\theta}} \nonumber\\
&= -i\sqrt{\eta} \db{\varphi}(\db{l})^\top \nabla \phi_{\eta}(\db{t}) e^{i\sqrt{\eta}\db{t}^{\top}\db{v} (\db{l}) }+\mathcal{O}(\eta^2).
\end{align} 
The orders of remainder terms are the same as in the univariate case. After that, we use another Taylor expansion on the exponential factor arising from the variable shift, $\exp(i\sqrt{\eta}\db{t}^{\top} \db{v} (\db{l}))$, at $\db{t} = \db{0}$,
\begin{equation}\label{taylorexp}
    \exp\big(i\sqrt{\eta}\,\db{t}^{\top} \db{v}(\db{l})\big)
= 1 + i\sqrt{\eta}\,\db{t}^{\top} \db{v}(\db{l})
- \frac{\eta}{2} \big( \db{t}^{\top} \db{v}(\db{l}) \big)^2
+ \mathcal{O}(\eta^{1.5}).
\end{equation}
We plug \eqref{vs1}-\eqref{taylorexp} into \eqref{updatedse}. By elementary calculation, the stationary equation becomes
\begin{equation}\label{pde_approx} 
-\db{t}^{\top} \Psi  \nabla \phi_{\eta}(\db{t})- \frac{\boldsymbol{t}^{\top}\E [\boldsymbol{v}^*\boldsymbol{v}^*{} ^{\top}] \boldsymbol{t} \phi_{\eta}(\db{t})} {2} = \tilde{\mathcal{R}}_{\eta}(\db{t}) ,  
\end{equation}
where $\db{v}^*  =  (v^*_1,\dots,v^*_d)^\top\in\R^d$ with $v^*_i=\tau_i -\ind_{X_i \leq \theta^*_i}$, $\Psi = \operatorname{diag}\big(f_{X_1}(\theta_1^{*}),\dots,f_{X_d}(\theta_d^{*})\big)$, and $\tilde{\mathcal{R}}_{\eta}(\db{t})=\mathcal{O} ( (-\log \eta)^3 \sqrt{\eta}) \rightarrow 0$ uniformly on bounded sets. Same as the proof in Section \ref{secproofmain}, all lower order terms cancel out, and
\begin{itemize}
    \item $\E [\db{v}^* (\db{v}^*)^{\top}]$ arises from the product of the zeroth-order term of the transition probability and the Hessian matrix of $\exp(i\sqrt{\eta}\db{t}^{\top} \db{v}(\db{l}))$.
    \item $\Psi$ emerges from the product of the gradient of the transition probability and the gradient of $\exp(i\sqrt{\eta}\db{t}^{\top} \db{v}(\db{l}))$.
\end{itemize}
Below we provide the derivation of $\E [\boldsymbol{v}^*\boldsymbol{v}^*{}^{\top} ]$, $\Psi$, and other lower order terms. For any fixed $\db{\theta}$, the family $\{E_{\db{l}}(\db{\theta}):\db{l}\in\{0,1\}^d\}$ is a partition of the entire sample space $\Omega$, i.e., the events are pairwise disjoint and their union is $\Omega$. Hence
\begin{equation}
\sum_{\db{l}\in\{0,1\}^d} P_{\db{l}}(\db{\theta})=1.
\label{eq:sumPs1}
\end{equation}
Fix $j\in\{1,\dots,d\}$. Consider the subfamily with $l_j=1$. The events
$\{E_{\db{l}}(\db{\theta}): l_j=1\}$ are pairwise disjoint and their union equals $\{X_j\le \theta_j\}$:
\begin{equation}
\bigcup_{\db{l}:\,l_j=1} E_{\db{l}}(\db{\theta})=\{X_j\le \theta_j\}.
\label{eq:union_sj1}
\end{equation}
So the sum of the probabilities is the marginal cumulative distribution,
\begin{equation}
\sum_{\db{l}:\,l_j=1} P_{\db{l}}(\db{\theta})=\Pbb(X_j\le \theta_j)=:F_{X_j}(\theta_j).
\label{eq:sumPs_sj1}
\end{equation}
\subsection*{Derivation of $\Psi$: product of the first order terms in both expansions}

%Notice that $\varphi_k(\db{l})$ can also be written as the integral of the joint density function. 
\iffalse
For fixed $s$ and $k$, define the $(d-1)$-dimensional region
\[
\mathcal{R}_{-k}(s;\theta^{*})
:=\Big\{x_{-k}\in\R^{d-1}:\ 
x_i\le \theta_i^{*}\ \text{if } s_i=1,\ 
x_i>\theta_i^{*}\ \text{if } s_i=0,\ \forall i\neq k
\Big\}.
\]
Then (Leibniz boundary differentiation) yields the cross-sectional formula
\begin{equation}
\varphi_k(s)
=(2s_k-1)\int_{\mathcal{R}_{-k}(s;\theta^{*})}
f_X(x_{-k},x_k=\theta_k^{*})\,dx_{-k}.
\label{eq:phi_cross}
\end{equation}
The sign $(2s_k-1)$ is positive when the constraint is $X_k\le \theta_k$
and negative when the constraint is $X_k> \theta_k$.
\fi
Let $A = \sum_{\db{l}\in\{0,1\}^d} \db{v}(\db{l})\db{\varphi}(\db{l})^\top$. To prove that $ A = -\operatorname{diag}\big(f_{X_1}(\theta_1^{*}),\dots,f_{X_d}(\theta_d^{*})\big)$, we proceed according to the following three steps:
\paragraph{Show $\sum_{\db{l}} \varphi_k(\db{l})=0$}
Differentiate \eqref{eq:sumPs1} with respect to $\theta_k$ at $\theta^{*}$:
\begin{equation}
\sum_{\db{l}\in\{0,1\}^d}\varphi_k(\db{l})
=\left.\frac{\partial}{\partial\theta_k}\sum_{\db{l}} P_{\db{l}}(\db{\theta})\right|_{\db{\theta}=\db{\theta}^{*}}
=\left.\frac{\partial}{\partial\theta_k}1\right|_{\db{\theta}=\db{\theta}^{*}}=0.
\label{eq:sumphi0}
\end{equation}
\paragraph{Get $A_{jk}$ entry-wisely}
For any $j,k$,
\begin{align}
A_{jk}
&=\sum_{\db{l}} v_j(\db{l})\varphi_k(\db{l})\nonumber\\
&=\sum_{\db{l}} (\tau_j-l_j)\varphi_k(\db{l})\nonumber\\
&=\tau_j\sum_{\db{l}} \varphi_k(\db{l})-\sum_{\db{l}} l_j\varphi_k(\db{l})
\nonumber\\
&=-\sum_{\db{l}} l_j\varphi_k(\db{l}),
\label{eq:Mjk_reduce}
\end{align}
where $\tau_j\sum_{\db{l}} \varphi_k(\db{l})=0$ due to \eqref{eq:sumphi0}.

\paragraph{Compute $\sum_{\db{l}} l_j\varphi_k(\db{l})$}
Note that
$$
\sum_{\db{l}} l_jP_{\db{l}}(\db{\theta})=\sum_{\db{l}:\,l_j=1} P_{\db{l}}(\db{\theta}).
$$
Differentiate w.r.t.\ $\theta_k$ at $\db{\theta}^{*}$ and use \eqref{eq:sumPs_sj1}:
\[
\sum_{\db{l}} l_j\varphi_k(\db{l})
=\left.\frac{\partial}{\partial\theta_k}\sum_{\db{l}:\,l_j=1}P_{\db{l}}(\db{\theta})\right|_{\db{\theta}=\db{\theta}^{*}}
=\left.\frac{\partial}{\partial\theta_k}F_{X_j}(\theta_j)\right|_{\db{\theta}=\db{\theta}^{*}}.
\]
Hence, for $k\neq j$, the derivative is zero; for $k=j$, it equals the marginal density
$f_{X_j}(\theta_j^{*})$ :
\begin{equation}
\sum_{\db{l}} l_j\varphi_k(\db{l})=
\begin{cases}
0, & k\neq j,\\
f_{X_j}(\theta_j^{*}), & k=j.
\end{cases}
\label{eq:sum_sj_phi}
\end{equation}
Combining \eqref{eq:Mjk_reduce} and \eqref{eq:sum_sj_phi},
\[
A_{jk}=
\begin{cases}
-\,f_{X_j}(\theta_j^{*}), & k=j,\\
0,& k\neq j.
\end{cases}
\]
Therefore
\begin{equation}
\sum_{\db{l}\in\{0,1\}^d} v(\db{l})\,\varphi(\db{l})^\top
=
-\operatorname{diag}\big(f_{X_1}(\theta_1^{*}),\dots,f_{X_d}(\theta_d^{*})\big)=-\Psi.
\end{equation}
\subsection*{Derivation of $\E [\boldsymbol{v}^*\boldsymbol{v}^*{}^{\top}]$}
%The $0$-th order term in Taylor's expansion of the transition probability is exactly $P_{\db{l}} (\db{\theta^*})$, for each $l \in \{0,1\}^d$. The $2$-nd order term in Taylor's expansion of $\exp(i\sqrt{\eta}\db{t}^{\top} \db{v} (\db{l}))$ yields a quadratic form of $-\eta \db{t}^{\top}\db{v} (\db{l}) \db{v} (\db{l})^{\top} \db{t}/2 $. 
By definition of $P_{\db{l}} (\db{\theta}^*)$ and $\db{v}(\db{l})$,
\begin{align*}
    \quad & \sum_{\db{l} \in \{0,1\}^d}P_{\db{l}}(\db{\theta}^*) \frac{-\db{t}^{\top}\db{v} (\db{l}) \db{v} (\db{l})^{\top} \db{t}}{2} \\
    & = \frac{- \db{t}^{\top} [\sum_{\db{l} \in \{0,1\}^d} P_{\db{l}}(\db{\theta}^*)\db{v} (\db{l}) \db{v} (\db{l})^{\top}] \db{t}}{2} \\
    & =\frac{- \db{t}^{\top} \E [\db{v}^* \db{v} ^*{}^{\top}] \db{t}}{2}. 
\end{align*}
By elementary calculations,
$$ (\E [\db{v}^* \db{v} ^*{}^{\top}])_{ij} = \mathbb{P}(X_i \leq \theta^*_i , X_j \leq \theta^*_j)-\tau_i\tau_j.$$
\subsection*{Derivation of lower order terms = 0}
The first observation is that the sum of the product of $0$-th order terms is exactly $1$ due to \eqref{eq:sumPs1}. This gives us $\phi_{\eta}(t) * 1$, which cancels out together with $\phi_{\eta}(t) $ on the left-hand side of \eqref{eqch}. 

We have also shown that $\sum_{\db{l}} \db{\varphi}(\db{l}) = \db{0}$, and similarly, the sum of $k$-th derivative of the transition kernel must be $0$. In other words, the contribution of the $0$-th order term of \eqref{taylorexp}, which is $1$, vanishes in the final equation. Notice that in equation \eqref{taylortran}, we omit the second order term in the Taylor expansion of the transition kernel because of this. The sum of the Hessian matrices over all $\db{l}$ yields $0$, and all remaining higher order interaction terms are absorbed into $\tilde{\mathcal{R}}_{\eta}(\db{t})$.

Then the only lower order term remaining is the product of the $0$-st order term in \eqref{taylortran} and the $1$-st order term in \eqref{taylorexp}. By the definition of $\db{v}^*$, it is clear that
$$ \sum_{\db{l}} P_{\db{l}}(\db{\theta}^*) \db{v}(\db{l}) = \E \db{v}^*=0. $$
Now we have justified the fact that all lower order terms cancel out.

Finally, we show that $\phi_\eta(\db{t})$ converges to the Gaussian characteristic function as $\eta \rightarrow 0$. Leveraging Theorem 1 in \cite{zhang_piecewise_2025} and Fatou's Lemma, we have that the second moments of $\db{\mathcal{P}}_{\eta}$ are uniformly bounded and the family $\{\db{\mathcal{P}}_{\eta}\}$ is tight (see also their discussion after Corollary 1). By Prokhorov's theorem and Lévy's continuity theorem, every subsequence $\eta_k \to 0$ has a further subsequence $\{\eta_{k_j}\}$ such that $\phi_{\eta_{k_j}}(\db{t}) \to \phi(\db{t})$ for some characteristic function $\phi(\db{t})$, and we also have the pointwise convergence of its gradient $\nabla \phi_{\eta_{k_j}}(\db{t}) \to \nabla\phi(\db{t})$. Taking the limit $\eta_{k_j}\rightarrow 0$ along the convergence subsequence in the perturbed PDE \eqref{pde_approx},
$$
-\db{t}^\top \Psi \nabla \phi(\db{t}) - \frac{1}{2} \db{t}^\top \E [\boldsymbol{v}^*\boldsymbol{v}^*{} ^{\top} ] \db{t} \phi(\db{t}) = 0, \quad \forall \db{t} \in \mathbb{R}^d.
$$
With the initial value condition $\phi(\db{0}) = 1$, this homogeneous PDE has the unique solution 
$$\phi(\db{t}) = \exp\!\left(-\frac{1}{2} \db{t}^\top V \db{t}\right)$$
where
\[
V = \int_0^{\infty} e^{-v \Psi^\top} \E[\boldsymbol{v}^* \boldsymbol{v}^{*\top}] e^{-v \Psi} \, dv.
\]
The matrix $V$ is also the unique solution to the Lyapunov equation $\Psi V + V \Psi^\top = \E [\boldsymbol{v}^*\boldsymbol{v}^*{} ^{\top} ]$. The closed form of $V$ as stated in the theorem can be obtained by elementary calculations. This solution $\phi(\db{t})$ is the characteristic function of $\mathcal{N}(0, V)$. Since every subsequence converges to $\phi(\db{t})$, we conclude the pointwise convergence $\phi_\eta(\db{t}) \to \phi(\db{t})$ for all $\db{t} \in \mathbb{R}^d$. It follows that $\db{\mathcal{P}}_{\eta}$ weakly converges to $\mathcal{N}(0, V)$ as $\eta \rightarrow 0$.
\end{proof}

\section*{Acknowledgment}
\addcontentsline{toc}{section}{Acknowledgment}
The authors would like to sincerely thank the Editor, the Associate Editor, and the two reviewers for their insightful comments which help us improve the clarity and the impact of the manuscript. Jiaqi Li's research is partially supported by NSF (Grant NSF/DMS-2515926). Likai Chen receives support from NSF (Grant NSF/DMS-2515927). Wei Biao Wu's research is partially supported by NSF (Grant NSF/DMS-2311249).

\bibliographystyle{IEEEtran}
\bibliography{IEEEabrv, reference}

\begin{IEEEbiographynophoto}{Ziyang Wei}
received the B.S. degree from Fudan University, Shanghai, China, in 2020. He is currently a Ph.D. Candidate in the Department of Statistics, The University of Chicago, Chicago, IL, USA. His research interests include stochastic optimization, online learning, and machine learning theory. His work focuses on the asymptotic theory, finite sample properties, and uncertainty quantification of stochastic gradient descent and online learning algorithms.
\end{IEEEbiographynophoto}
\begin{IEEEbiographynophoto}{Jiaqi Li}
received the Ph.D. degree in Statistics from Washington University in St. Louis in 2024. She is a William H. Kruskal Instructor in the Department of Statistics at the University of Chicago, and is joining the Department of Statistics at Rice University as an Assistant Professor in July 2026. Her research interests include machine learning theory, time series, high-dimensional data and neuroimaging.
\end{IEEEbiographynophoto}
\begin{IEEEbiographynophoto}{Likai Chen}
received the Ph.D. degree in statistics from the University of
Chicago in 2018. She is currently an Associate Professor with the Department
of Statistics and Data Science, Washington University in St. Louis. Her
research interests include time series analysis, high-dimensional statistics, and statistical learning theory.
\end{IEEEbiographynophoto}
\begin{IEEEbiographynophoto}{Wei Biao Wu}
received the B.S. degree from Fudan University, Shanghai, China, in 1997, and the Ph.D. degree in statistics from the University of Michigan, Ann Arbor, MI, USA, in 2001.

He is currently a Professor with the Department of Statistics, The University of Chicago, Chicago, IL, USA. His research interests include probability theory, time series analysis, high-dimensional statistical inference, and machine learning. His work focuses on the development of asymptotic theory for complex dependence structures, functional dependence measures, and the statistical analysis of stochastic gradient descent and online learning algorithms.

He was the recipient of the National Science Foundation CAREER Award in 2004, the Tjalling C. Koopmans Econometric Theory Prize in 2009, and the Humboldt Research Award from the Alexander von Humboldt Foundation in 2019. He has served as an Associate Editor for several journals, including the Annals of Statistics, Bernoulli and Journal of the American Statistical Association.
\end{IEEEbiographynophoto}

\end{document}